\PassOptionsToPackage{hyphens}{url}
\documentclass[11pt,letterpaper,logo,onecolumn]{hy_llmeval_style}
\renewcommand{\today}{2026-07-31}

\usepackage[authoryear,sort&compress,round]{natbib}
\usepackage{booktabs}
\usepackage{multirow}
\usepackage{amsmath}
\usepackage{amssymb}
\usepackage{pifont}
\usepackage{graphicx}
\usepackage{threeparttable}
\usepackage{subcaption}
\usepackage[normalem]{ulem}
\usepackage[most,breakable,skins]{tcolorbox}
\tcbuselibrary{skins,breakable}
\usepackage{microtype}
\usepackage{tikz}
\usetikzlibrary{positioning,calc,backgrounds}
\usepackage{tabularx}
\usepackage{float}
\usepackage{placeins}
\usepackage{iftex}
\ifPDFTeX
  \usepackage{CJKutf8}
\else
  \usepackage{xeCJK}
  \xeCJKDeclareCharClass{CJK}{"2460 -> "2473}
\fi
\let\cite\citep
\hypersetup{
  citecolor=gblue9,
  linkcolor=gblue9,
  urlcolor=gblue9
}

\newcommand{\bench}{\textsc{Hy-MultiTurn}}
\newcommand{\shortcite}[1]{\cite{#1}}

\title{Hy-MultiTurn: A Six-Dimensional Benchmark for Deep\\Multi-Turn Dialogue Understanding}

\author{
  Eileen Ye\textsuperscript{\rm 1*},
  Jiawen Tao\textsuperscript{\rm 1,2*},
  Yaoming Li\textsuperscript{\rm 2*},
  Chenxu Liu\textsuperscript{\rm 1},
  Wenhan Yu\textsuperscript{\rm 2},\\
  Yaxin Fan\textsuperscript{\rm 1},
  Xiaokun Yuan\textsuperscript{\rm 1,2},
  Mengzhou Wu\textsuperscript{\rm 1},
  Yanbing Jiang\textsuperscript{\rm 2\dag},
  Maxm Pan\textsuperscript{\rm 1\dag}\\
  \vspace{0.3cm}
  \normalsize
  \textsuperscript{1}Hunyuan Team, Tencent\\
  \textsuperscript{2}Peking University\\
  \texttt{\normalsize \{eileenye,jadentao\}@tencent.com}
}

\begin{abstract}
Long-running multi-turn interactions with chatbots and agents are now common, and a correct response often depends on remembering earlier details, tracking later revisions, identifying intended objects or referents, and withholding action when required conditions are unmet. Existing multi-turn benchmarks typically cover short exchanges and do not fully evaluate these capabilities in long multi-turn interactions, particularly in Chinese, while offering limited insight into how and why models fail. To address these limitations, we analyze real chatbot failures to identify six recurring mechanisms and use them to define six controlled evaluation modes in \bench{}, a Chinese benchmark for deep multi-turn dialogue understanding. The six modes evaluate constraint memory, precise execution, constraint synthesis, object localization, action suppression, and reference resolution. Across the six modes, we construct 209 controlled tasks spanning 12--76 turns, with dialogue length, irrelevant-topic distraction, and colloquial phrasing adding further difficulty. Evaluation of 22 frontier model configurations shows that \bench{} is broadly challenging, as even GPT-5.5, the strongest overall configuration, satisfies all requirements in only 41.1\% of responses and no model performs best in all six modes.
\end{abstract}

\begin{document}
\raggedbottom
\fancyhead[C]{\footerfont Hy-MultiTurn: A Six-Dimensional Benchmark for Deep Multi-Turn Dialogue Understanding}

\begingroup
  \renewcommand\thefootnote{}
  \footnote{\hspace{-1.8em}\textsuperscript{*}Equal contribution.\\
            \textsuperscript{\dag}Corresponding author.}
\endgroup
\maketitle

\section{Introduction}

Chatbots and agents increasingly support long-running multi-turn interactions in customer-service workflows, collaborative planning, and document iteration. Rather than issuing a self-contained request, users build up a task over many turns by introducing constraints casually, revising earlier decisions, and referring back to previously mentioned entities. To produce a correct final response, a model must maintain and update the dialogue state throughout the interaction, including which constraints remain active, which values have been revised, which object or referent the user intends, and whether the conditions for taking action have been met. We refer to the capability required in these long-running interactions as deep multi-turn understanding. It requires more than producing a response that is fluent or locally plausible; the response must remain consistent with the valid state of the conversation as a whole.

Fluent responses may still contain substantive errors. Models may forget an early low-salience constraint, reuse an overridden value, select the wrong object after its attributes change, resolve an ambiguous pronoun to a recent distractor, or produce an executable plan without authorization. Consider a user who casually mentions owning a cat early in the conversation and, 40+ turns later, asks for flower recommendations for the home. A model that overlooks this pet-safety fact may still recommend lilies, which are toxic to cats. The recommendation may appear relevant and plausible, yet be unusable because it violates one decisive constraint.

Existing public evaluations address individual components of this challenge but do not assess them jointly. MT-Bench \cite{zheng2023judging} focuses on general response quality in two-turn exchanges; Multi-IF \cite{he2024multiif} and MT-Eval \cite{kwan2024mteval} extend instruction following or recollection across a few turns; and long-context benchmarks such as RULER \cite{hsieh2024ruler}, BABILong \cite{kuratov2024babilong}, and LongBench \cite{bai2024longbench} emphasize retrieval and reasoning over static documents. Taken together, these settings do not fully test whether a final response incorporates information introduced earlier, reflects later revisions, associates attributes with the correct entities, resolves indirect references, and withholds action when required conditions are not met. The central challenge is therefore not context length alone, but whether the model can maintain the current valid dialogue state and use it to produce a fully correct final response.

\begin{table*}[t]
\centering
{\small
\setlength{\tabcolsep}{2.0pt}
\newcommand{\yes}{{\color{green!70!black}\ding{51}}}
\newcommand{\no}{{\color{red!70!black}\ding{55}}}
\newcommand{\pmark}{{\color{orange!80!black}$\boldsymbol{\sim}$}}
\resizebox{\textwidth}{!}{%
\begin{tabular}{@{}l c c c c c c c c@{}}
\toprule
& \textbf{MT-Bench} & \textbf{Multi-IF} & \textbf{MT-Eval} & \textbf{RULER} & \textbf{LongMemEval} & \textbf{C3} & \textbf{MultiCh.} & \cellcolor{blue!8}\textbf{Hy-MT} \\
& Zheng'23 & He'24 & Kwan'24 & Hsieh'24 & Wu'25 & Ma'25 & Deshpande'25 & \cellcolor{blue!8}Ours \\
\midrule
\multicolumn{9}{l}{\textit{Basic properties (shared across benchmarks)}} \\
Multi-turn interaction & \yes & \yes & \yes & \no & \yes & \pmark & \yes & \cellcolor{blue!8}\yes \\
Chinese language & \no & \pmark & \no & \no & \no & \yes & \no & \cellcolor{blue!8}\yes \\
Objective/verifiable scoring & \no & \yes & \no & \yes & \yes & \pmark & \yes & \cellcolor{blue!8}\yes \\
Real-world failure grounding & \no & \no & \pmark & \no & \no & \pmark & \pmark & \cellcolor{blue!8}\yes \\
Instruction following & \yes & \yes & \yes & \no & \no & \no & \yes & \cellcolor{blue!8}\yes \\
Information retrieval & \no & \no & \pmark & \yes & \yes & \no & \pmark & \cellcolor{blue!8}\yes \\
\midrule
\multicolumn{9}{l}{\textit{Deep multi-turn capabilities (differentiating dimensions)}} \\
$>$20 dialogue turns & \no & \no & \no & N/A & \pmark & \no & \no & \cellcolor{blue!8}\yes \\
Cross-turn state tracking & \no & \pmark & \pmark & \pmark & \pmark & \pmark & \yes & \cellcolor{blue!8}\yes \\
Constraint update \& override & \no & \no & \no & \no & \pmark & \no & \pmark & \cellcolor{blue!8}\yes \\
Action suppression & \no & \no & \no & \no & \no & \no & \no & \cellcolor{blue!8}\yes \\
Long-range reference resolution & \no & \no & \no & \pmark & \no & \pmark & \no & \cellcolor{blue!8}\yes \\
Item-level failure diagnostics & \no & \pmark & \no & \no & \no & \no & \pmark & \cellcolor{blue!8}\yes \\
\bottomrule
\end{tabular}
}
}
\caption{Comparison with representative multi-turn and long-context benchmarks.
{\color{green!70!black}\ding{51}} = benchmark-wide or broad coverage;
{\color{orange!80!black}$\boldsymbol{\sim}$} = limited subset, task-specific, or proxy coverage;
{\color{red!70!black}\ding{55}} = not covered.
MultiCh.\ = MultiChallenge.}
\label{tab:comparison}
\end{table*}

Evaluating directly on returned production conversations looks natural, since that is where failures occur. But raw logs are uncontrollable, non-reproducible, and entangled in cause---hard to attribute cleanly---precisely the limitation of production bad-case test sets. We therefore use real data only to reverse-engineer six recurring mechanisms, then regenerate all benchmark dialogues under controlled conditions with no real user content. Difficulty is controlled separately through dialogue length, irrelevant-topic distraction, and colloquial phrasing. Each task is anchored to a traceable ground truth, stress-tested with irrelevant context that cannot change the answer, and decomposed into verifiable scoring items. Answers and scoring are frozen before distractors are added, after which candidates are blind-selected for clean, discriminative failures. This separates the target failure mechanism from generic difficulty and identifies \textit{where} and \textit{why} a response fails rather than assigning one holistic score. We detail mechanism induction and the held-out coverage audit in Section~\ref{sec:mechanism-induction}, and the task-construction pipeline in Section~\ref{sec:construction-pipeline}.

\begin{figure*}[t]
\centering
\includegraphics[width=0.90\textwidth]{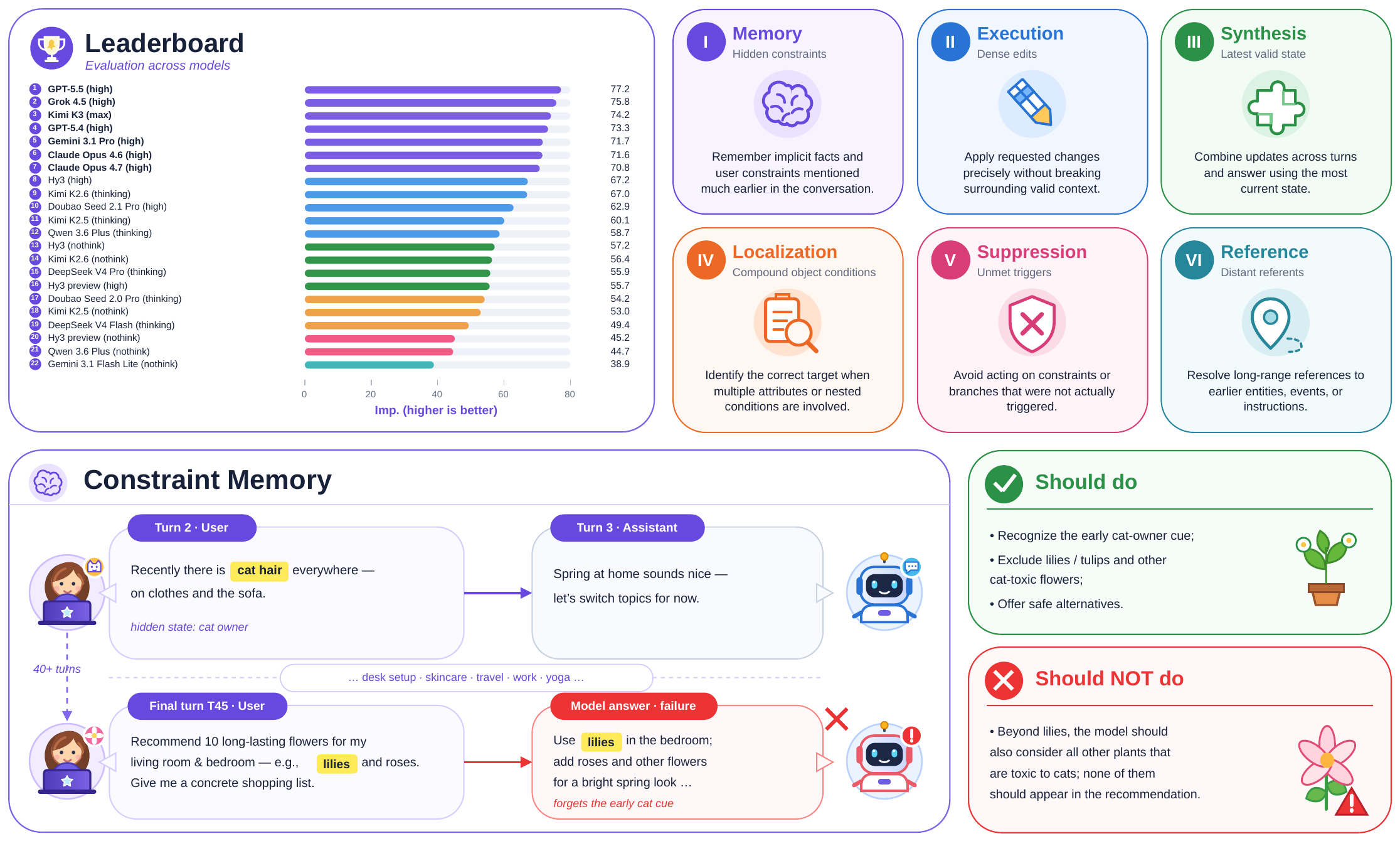}
\caption{\bench{} overview: the 22-model leaderboard, six evaluation dimensions, and a Mode~I example requiring an early pet-safety constraint to be applied after 40+ turns.}
\label{fig:overview}
\end{figure*}

To address these limitations, we present \bench{}, a Chinese benchmark for deep multi-turn dialogue understanding. Its main contributions are threefold.

\begin{enumerate}
\item A \textbf{Chinese long-horizon multi-turn benchmark} that turns broad dialogue failure into six separately measurable dimensions: constraint memory, precise execution, constraint synthesis, object localization, action suppression, and reference resolution. It covers 209 tasks of 12--76 turns.
\item A \textbf{failure-mechanism-first construction and scoring methodology} that instantiates each mechanism as a controlled evaluation mode and anchors every task to a traceable ground truth. The pipeline separates mode-specific challenges from dialogue length, irrelevant-topic distraction, and colloquial phrasing, with answers and scoring criteria fixed before distractors are added. Responses are evaluated against 6,317 item-level criteria.
\item A \textbf{three-roll evaluation of 22 frontier model configurations} using weighted importance to measure correctness across requirements according to their importance and strict accuracy to determine whether all requirements are satisfied. Together, these metrics better reflect the user experience and practical usability of multi-turn responses. Partial credit may hide errors or omissions that prevent direct use, whereas strict accuracy exposes them. GPT-5.5 leads on both metrics, while different models lead across the six modes.
\end{enumerate}

\section{Related Work}

\subsection{Multi-Turn Dialogue Evaluation}

MT-Bench \cite{zheng2023judging} established LLM-as-judge with 80 two-turn questions, but frontier models now exceed 9/10 (cf.\ Arena-Hard, WildBench, AlpacaEval~\cite{li2025arenahard,lin2025wildbench,dubois2024alpacaeval}). MT-Bench-101 \cite{bai2024mtbench101} adds 13 tasks including context memory, but lacks systematic difficulty design and deterministic scoring. MT-Eval \cite{kwan2024mteval} covers four interaction types in 6.96-turn dialogues but lacks mechanism-specific diagnostics and controlled distractor design.

Multi-IF \cite{he2024multiif} measures instruction forgetting over three turns in eight languages (best model: 87.7\% to 70.7\%); EvolIF \cite{jia2025evolif} uses Flow Theory to escalate difficulty, with GPT-5 leading at 66.4\% stability; and Laban et al.'s Sharded Simulation \cite{laban2025lost} finds 39\% average degradation and four systematic failure modes. These extend single-turn IFEval \cite{zhou2023ifeval} through fine-grained constraints (FollowBench \cite{jiang2024followbench}, InFoBench \cite{qin2024infobench}, ComplexBench \cite{wen2024complexbench}) and multi-turn interaction (Parrot \cite{sun2024parrot}); Chinese work includes CMT-Eval \cite{tian2025cmteval}, FB-Bench \cite{li2025fbbench}, and AlignBench \cite{liu2024alignbench}.

Recent work targets more realistic and structured interaction: MultiChallenge \cite{deshpande2025multichallenge} identifies four recurring challenges with instance-level rubrics; StructFlowBench \cite{li2025structflowbench} models inter-turn structural relations; BotChat \cite{duan2024botchat} and MathChat \cite{liang2024mathchat} stress long or domain-specific dialogues; FairMT-Bench \cite{fan2025fairmt} studies fairness accumulation; and MINT and AgentBench~\cite{wang2024mint,liu2024agentbench} evaluate tool-using agents.

\bench{} complements these benchmarks with fixed Chinese-native dialogues of 12--76 turns, controlled distractors, and mechanism-specific scoring. Relative to single-prompt IFEval \cite{zhou2023ifeval} and FollowBench \cite{jiang2024followbench}, Mode~II checks dozens of accumulated and revised constraints, and the suite also tests low-salience memory, state synthesis, and action suppression.

\subsection{Long-Context and Reference Resolution}

LongBench~\cite{bai2024longbench,bai2024longbench2}, $\infty$Bench \cite{zhang2024infinitebench}, BABILong \cite{kuratov2024babilong}, and RULER \cite{hsieh2024ruler} test long-document and NIAH-style retrieval, while LongMemEval \cite{wu2025longmemeval} and MemoryBank \cite{zhong2024memorybank} cover long-term memory. These use static documents or discrete sessions rather than one continuously revised dialogue; Mode~IV instead needs \emph{compound} object conditions.

C3 \cite{ma2025c3} covers bilingual spoken dialogues up to 16 turns; Hardt \shortcite{hardt2023ellipsis} finds failures in ellipsis-dependent reasoning beyond simple structures; IHEval \cite{zhang2025iheval} tests instruction hierarchy (best open-source model: 48\%); and MultiTurnInstruct \cite{han2025multiturninstruct} tests entangled instructions but not action suppression.

\bench{} fills two gaps: knowing when \emph{not} to act under unmet preconditions, and recovering a referent introduced 20+ turns earlier despite a salient recent distractor. Mode~VI tests 20+ turn reference resolution with controlled recent distractors, jointly requiring linguistic competence and long-range tracking. Table~\ref{tab:comparison} situates \bench{} against these lines of work: it is Chinese-native, spans 12--76 turns, and jointly covers cross-turn state tracking, constraint update, action suppression, long-range reference, and item-level diagnostics.

\section{Benchmark Design and Construction}

\subsection{Design Philosophy}

\bench{} follows three principles. \textbf{Failure-mode-first design} defines each dimension around an observable cognitive failure rather than holistic dialogue quality. \textbf{State-based difficulty} comes from maintaining a unique valid dialogue state across turns rather than from underspecified tasks; each of the six dimensions isolates a distinct failure mechanism, while task difficulty is varied through dialogue length, irrelevant-topic distraction, and colloquial phrasing. \textbf{Item-level accountability} decomposes each task into independently verifiable weighted items, with gates for core conclusions where needed, thereby locating which capability fails while preserving a unique answer derivable from the dialogue.

\subsection{Six Cognitive Dimensions}
\label{sec:mechanism-induction}

We induce six recurring failure mechanisms from desensitized returned multi-turn failures of a large-scale Chinese chatbot and treat each as a capability dimension with a corresponding evaluation mode. The same review identified three controllable factors recurring across mechanisms: longer dialogue histories increase state-retention demands, irrelevant topic shifts compete for attention, and colloquial phrasing leaves constraints and references less explicit. We vary these factors independently rather than treat them as additional modes. We then validate their coverage on a held-out sample drawn from a pre-filtered pool of likely problematic conversations. Of 1{,}875 labeled dialogues in that audit, 1{,}354 (72.2\%) contain a substantive model problem; among these problem cases, 643 (47.5\%) match one of the six mechanisms below, while the rest are mainly hallucinations, tool or generation failures, domain-competence errors, or safety/refusal issues rather than multi-turn capability failures. Most multi-turn capability errors in the audit fall into this six-way taxonomy (Supplementary Section~C). Following this one-to-one mapping, Mode~I retains an early low-salience constraint; Mode~II executes accumulated and revised requirements; Mode~III tracks the latest valid values and their dependencies; Mode~IV selects objects under compound attribute conditions; Mode~V withholds action when formal preconditions are unmet; and Mode~VI resolves distant references against recent distractors. Together they separate memory, state-update, binding, and behavioral-control failures that a holistic dialogue-quality score would otherwise conflate. The final benchmark contains 34 Mode~I tasks and 35 in each remaining mode (209 total). Across modes, tasks span 12--76 turns and contain 3--176 scoring items; detailed per-mode statistics appear in Supplementary Section~B.

\subsection{Scoring Framework}

Each task is decomposed into independently checkable hit/miss items, aggregated per task and then averaged across tasks. This decomposition localizes which requirement failed instead of assigning one holistic quality score. Mode~II uses a deterministic checker; the other modes use LLM judges. Weighted \textbf{rubric items} cover verifiable details, while Modes~I, V, and VI additionally use a binary \textbf{gate item} for the core conclusion.

The same judgments yield two complementary metrics. For rubric weights $w_i$, hit set $H$, and $m$ misses, \textit{importance} $=\sum_{i\in H}w_i/\sum_iw_i$ measures how much weighted content is correct; \textit{strict} $=\mathbb{1}[m{=}0]$ requires every item, indicating whether a response is usable verbatim. Gate failure sets both to zero regardless of rubric credit. \textit{Importance} is the \textbf{primary ranking metric}: its partial credit remains comparable across tasks with different rubric granularity and best supports diagnosis and ranking. \textit{Strict} is the \textbf{primary usability metric}, because in professional settings one inconsistent amount, dropped safety constraint, or stale value can require human correction despite otherwise strong coverage.

Gates capture \textbf{qualitative failures that partial credit cannot offset}: recommending cat-toxic lilies (I), emitting an unauthorized quarantine table (V), or resolving the wrong entity (VI) invalidates otherwise correct details. Modes~II--IV instead measure cumulative success over format and numeric constraints, fields reconciled to their latest state, or object--attribute bindings. Because a gate would equate one miss with twenty and discard diagnostic signal, gates represent binary failure mechanisms while pure rubrics preserve graded ones.

\subsection{Unified Construction Pipeline}
\label{sec:construction-pipeline}

Combining the six mechanisms with three cross-cutting factors, we construct controlled tasks that separate \emph{what} fails from what makes the task difficult. Real conversations are used only for mechanism induction; all 209 benchmark dialogues are rewritten or synthesized and contain no real user content. This preserves production-grounded failure patterns while keeping each mechanism controlled and reproducible.

\begin{figure*}[t]
\centering
\includegraphics[width=0.84\textwidth]{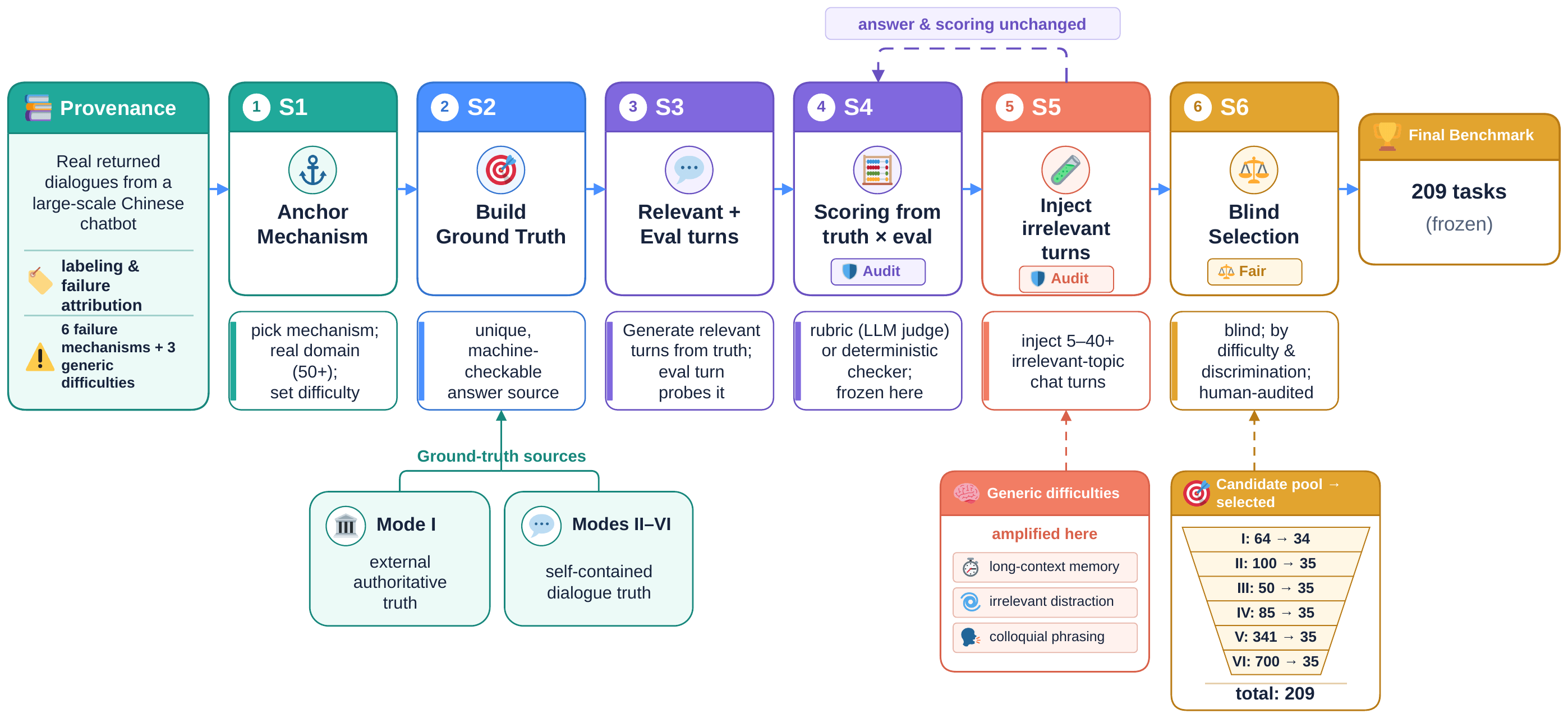}
\caption{Unified six-stage construction pipeline shared across dimensions: mechanism anchoring, traceable ground truth, dialogue construction, scoring construction, distractor injection, and blind selection.}
\label{fig:pipeline}
\vspace{-0.4cm}
\end{figure*}

All tasks follow the six stages in Fig.~\ref{fig:pipeline}. We first anchor one target mechanism and choose a matched scenario, controlling difficulty through dialogue length, irrelevant content, and the distance from key information to the final query (S1). We then construct a unique, traceable ground truth (S2). For Mode~I, we retrieve authoritative legal, medical, and toxicological sources and record each fact with its source and applicable boundary. For Modes~II--VI, the ground truth is fixed by the task's self-contained state, represented respectively by a constraint graph, state-update chain, object--attribute registry, trigger-condition boundary, or explicit referent--distractor record. Each representation records the current valid state and relevant alternatives so that a single correct answer can be derived. Relevant information is distributed across the dialogue, and the final evaluation turn is written in natural business language, without mode labels, so that it cannot be answered correctly without recovering and integrating that state (S3).

Scoring items are derived jointly from the evaluation request, which determines what is tested, and the ground truth, which fixes the correct value (S4). Items are atomic and audited against truth entries for omissions or conflicts. In judge-scored modes, criterion-specific clarification notes distinguish required facts from acceptable surface forms, reducing stylistic false negatives. Mode~II uses a deterministic checker validated with full-score fixtures, targeted mutations, and false-negative audits; Modes~I, V, and VI add gates for errors that invalidate the response. Other modes retain graded item-level credit. Crucially, we freeze the ground truth, answer, and scoring items---including any gate---before introducing distractors or model-based filtering.

We then inject 5--40+ irrelevant turns to increase memory load and control information distance (S5). A boundary audit checks that these turns neither change nor reveal the frozen answer. For example, in a Mode~I livestreaming task, an early statement that the user is a seventh-grade student fixes the gate answer. Any injected turn that would change this answer, such as one implying adulthood, is rejected. Finally, tasks are selected from larger candidate pools using cross-model difficulty and discrimination while hiding model identity and without targeting any specific model (S6). Human spot checks remove cases where apparent failures arise from scoring artifacts rather than model behavior. Full mode-specific ground-truth structures, evaluation-turn templates, checker audits, and filtering procedures appear in Supplementary Sections~B and~G.

The selection pool contains 1{,}340 candidates across Modes~I--VI (64/100/50/85/341/700), of which 209 are retained. Blind cross-model screening uses difficulty, discrimination, and scoring validity rather than any target model. Evidence is mechanism-specific: Mode~II requires checker fixtures and false-negative audits; Modes~I, V, and VI receive gate review; Mode~IV requires selector proofs; and judge-based modes are audited for item completeness. We exclude ambiguous ground truth, fragile extraction, saturation, weak distractors, and non-diagnostic universal failures (Supplementary Section~G).

\subsection{Worked Example}

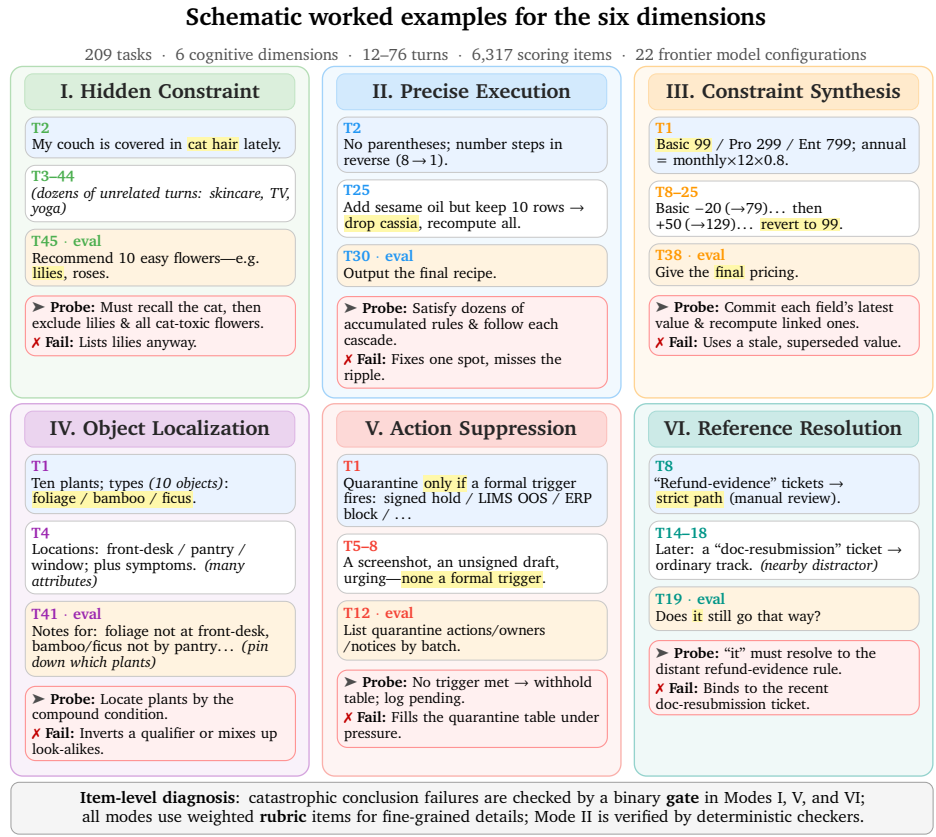
\begin{figure}[H]
\noindent\hspace*{-0.055\textwidth}\makebox[\textwidth][c]{%
  \resizebox{0.86\textwidth}{!}{\begingroup

\providecommand{\bench}{\textsc{Hy-MultiTurn}}

\newcommand{\xmark}{{\color{red!76!black}\ding{55}}}
\newcommand{\probeicon}{{\color{black!70}\ding{228}}}
\newcommand{\rolelab}[1]{{\color{black!45}\fontsize{5.9}{6.3}\selectfont\textbf{#1:}}}
\newcommand{\turn}[2]{{\color{#2}\fontsize{7.4}{8.0}\selectfont\bfseries #1}\\[0.5pt]}
\newcommand{\turntest}[2]{{\color{#2}\fontsize{7.4}{8.0}\selectfont\bfseries #1 $\cdot$ eval}\\[0.5pt]}
\newcommand{\hl}[1]{%
  \begingroup
  \setlength{\fboxsep}{0.45pt}%
  \colorbox{yellow!42}{#1}%
  \endgroup
}
\newcommand{\mechanism}[1]{\textcolor{black!62}{\textbf{#1}}}
\newcommand{\correct}[1]{\textcolor{blue!65!black}{\textbf{#1}}}
\newcommand{\distract}[1]{\textcolor{orange!82!black}{\textbf{#1}}}

\definecolor{cI}{HTML}{4CAF50}
\definecolor{cII}{HTML}{2196F3}
\definecolor{cIII}{HTML}{FF9800}
\definecolor{cIV}{HTML}{9C27B0}
\definecolor{cV}{HTML}{F44336}
\definecolor{cVI}{HTML}{009688}

\definecolor{userblue}{HTML}{EAF3FF}
\definecolor{traporange}{HTML}{FFF3E0}
\definecolor{botred}{HTML}{FFF0F0}

\newcommand{\cardheight}{6.72}
\newcommand{\firstBubbleY}{0.86}
\newcommand{\bubbleYsep}{3.0pt}
\newcommand{\probeYsep}{3.0pt}

\newcommand{\chatcard}[9]{%
  \begin{scope}[shift={(#1,#2)}]

    \draw[
      rounded corners=6pt,
      draw=#3!42,
      line width=0.65pt,
      fill=#3!6
    ] (0,0) rectangle (5.12,-\cardheight);

    \node[
      hdr,
      fill=#3!20,
      draw=#3!35,
      anchor=north
    ] (hd) at (2.56,-0.12) {#4};

    \node[
      ub,
      fill=userblue,
      anchor=north west
    ] (b1) at (0.245,-\firstBubbleY) {#5};

    \node[
      ub,
      fill=white,
      anchor=north west
    ] (b2) at ($(b1.south west)+(0,-0.112)$) {#6};

    \node[
      ub,
      fill=traporange,
      anchor=north west
    ] (b3) at ($(b2.south west)+(0,-0.112)$) {#7};

    \node[
      bb,
      anchor=north west
    ] (rb) at ($(b3.south west)+(0,-0.154)$)
      {\probeicon\ \textbf{Probe:}\ #8\\[1.5pt]\xmark\ \textbf{Fail:}\ #9};

  \end{scope}
}

\hyphenpenalty=10000
\exhyphenpenalty=10000

\begin{tikzpicture}[
  font=\sffamily,
  hdr/.style={
    rounded corners=4pt,
    line width=0.4pt,
    text width=4.40cm,
    inner xsep=3.3pt,
    minimum height=0.50cm,
    text height=8.2pt,
    text depth=2.0pt,
    font=\small\bfseries,
    align=center,
    text=black!85
  },
  ub/.style={
    rounded corners=5pt,
    draw=black!22,
    line width=0.45pt,
    align=left,
    font=\fontsize{7.25}{7.65}\selectfont,
    text width=4.40cm,
    inner xsep=3.3pt,
    inner ysep=\bubbleYsep,
    minimum height=0.54cm
  },
  bb/.style={
    rounded corners=5pt,
    draw=red!42,
    fill=botred,
    line width=0.55pt,
    align=left,
    font=\fontsize{7.15}{7.75}\selectfont,
    text width=4.40cm,
    inner xsep=3.3pt,
    inner ysep=\probeYsep,
    minimum height=0.60cm
  },
  foot/.style={
    font=\fontsize{6.8}{7.4}\selectfont\bfseries,
    align=left,
    text width=4.55cm
  },
]

\node[
  font=\large\bfseries
] (title) at (7.98,0.26)
  {Schematic worked examples for the six dimensions};

\node[
  font=\scriptsize,
  text=black!68,
  below=0.7pt of title
]
  {209 tasks \;$\cdot$\; 6 cognitive dimensions \;$\cdot$\; 12--76 turns \;$\cdot$\; 6{,}317 scoring items \;$\cdot$\; 22 frontier model configurations};

\renewcommand{\cardheight}{5.68}
\renewcommand{\bubbleYsep}{2.55pt}
\renewcommand{\probeYsep}{2.60pt}

\chatcard{0}{-0.56}{cI}
  {I. Hidden Constraint}
  {\turn{T2}{cI} My couch is covered in \hl{cat hair} lately.}
  {\turn{T3--44}{cI} \textit{(dozens of unrelated turns: skincare, TV, yoga)}}
  {\turntest{T45}{cI} Recommend 10 easy flowers---e.g.\ \hl{lilies}, roses.}
  {Must recall the cat, then exclude lilies \& all cat-toxic flowers.}
  {Lists lilies anyway.}

\chatcard{5.35}{-0.56}{cII}
  {II. Precise Execution}
  {\turn{T2}{cII} No parentheses; number steps in reverse (8\,$\to$\,1).}
  {\turn{T25}{cII} Add sesame oil but keep 10 rows $\to$ \hl{drop cassia}, recompute all.}
  {\turntest{T30}{cII} Output the final recipe.}
  {Satisfy dozens of accumulated rules \& follow each cascade.}
  {Fixes one spot, misses the ripple.}

\chatcard{10.70}{-0.56}{cIII}
  {III. Constraint Synthesis}
  {\turn{T1}{cIII} \hl{Basic 99} / Pro 299 / Ent 799; annual = monthly$\times$12$\times$0.8.}
  {\turn{T8--25}{cIII} Basic $-$20\,($\to$79)\dots\ then $+$50\,($\to$129)\dots\ \hl{revert to 99}.}
  {\turntest{T38}{cIII} Give the \hl{final} pricing.}
  {Commit each field's latest value \& recompute linked ones.}
  {Uses a stale, superseded value.}

\renewcommand{\cardheight}{6.38}
\renewcommand{\bubbleYsep}{3.0pt}
\renewcommand{\probeYsep}{3.0pt}

\chatcard{0}{-6.34}{cIV}
  {IV. Object Localization}
  {\turn{T1}{cIV} Ten plants; types \textit{(10 objects)}:\\ \hl{foliage / bamboo / ficus}.}
  {\turn{T4}{cIV} Locations: front-desk / pantry / window; plus symptoms. \textit{(many attributes)}}
  {\turntest{T41}{cIV} Notes for: foliage not at front-desk, bamboo/ficus not by pantry\dots\ \textit{(pin down which plants)}}
  {Locate plants by the compound condition.}
  {Inverts a qualifier or mixes up look-alikes.}

\chatcard{5.35}{-6.34}{cV}
  {V. Action Suppression}
  {\turn{T1}{cV} Quarantine \hl{only if} a formal trigger fires: signed hold / LIMS OOS / ERP block / \dots}
  {\turn{T5--8}{cV} A screenshot, an unsigned draft, urging---\hl{none a formal trigger}.}
  {\turntest{T12}{cV} List quarantine actions/owners\\/notices by batch.}
  {No trigger met $\to$ withhold table; log pending.}
  {Fills the quarantine table under pressure.}

\chatcard{10.70}{-6.34}{cVI}
  {VI. Reference Resolution}
  {\turn{T8}{cVI} ``Refund-evidence'' tickets $\to$ \hl{strict path} (manual review).}
  {\turn{T14--18}{cVI} Later: a ``doc-resubmission'' ticket $\to$ ordinary track. \textit{(nearby distractor)}}
  {\turntest{T19}{cVI} Does \hl{it} still go that way?}
  {``it'' must resolve to the distant refund-evidence rule.}
  {Binds to the recent doc-resubmission ticket.}

\node[
  draw=black!38,
  rounded corners=4pt,
  fill=black!4,
  align=center,
  font=\scriptsize,
  text width=15.55cm,
  inner xsep=3.8pt,
  inner ysep=4.0pt,
  anchor=north west
] at (0,-12.82)
  {\textbf{Item-level diagnosis}: catastrophic conclusion failures are checked by a binary \textbf{gate} in Modes I, V, and VI; all modes use weighted \textbf{rubric} items for fine-grained details; Mode II is verified by deterministic checkers.};

\end{tikzpicture}

\endgroup}%
}
\caption{Representative task patterns for the six evaluation modes. Each example shows the relevant earlier information, later distractor or pressure, final test, and a typical failure. Appendix~\ref{app:worked-examples} provides the complete worked examples.}
\label{fig:worked-examples}
\end{figure}

Figure~\ref{fig:worked-examples} summarizes representative tasks for all six modes. Each example shows the relevant earlier information, later distractor or pressure, final test, and a typical failure. In Mode~V, a pharmaceutical-quality dialogue defines formal triggers for quarantining a batch, but later provides only insufficient signals before requesting an executable plan. The correct response records the known and pending conditions without producing the plan; acting before a trigger is met is a gate failure. Appendix~\ref{app:worked-examples} provides the complete examples and analyses for all six modes.

\section{Experiments}

\newcommand{\renderleaderboardtable}{%
\begin{table*}[t]
\centering
{\small
\setlength{\tabcolsep}{4.8pt}
\begin{tabular}{@{}r l @{\hspace{10pt}} cc @{\hspace{8pt}} cccccc@{}}
\toprule
& & \multicolumn{2}{c}{\textbf{Overall}} & \multicolumn{6}{c}{\textbf{Per-dimension importance}} \\
\cmidrule(lr){3-4}\cmidrule(l){5-10}
\textbf{\#} & \textbf{Model} & \textbf{Imp.} & \textbf{Str.} & \textbf{I} & \textbf{II} & \textbf{III} & \textbf{IV} & \textbf{V} & \textbf{VI} \\
\midrule
1 & GPT-5.5 (high) & \textbf{77.2} & \textbf{41.1} & \textbf{71.2} & 86.2 & 84.9 & 90.7 & 60.9 & 69.0 \\
2 & Grok 4.5 (high) & 75.8 & 39.9 & 48.9 & 79.5 & \textbf{90.0} & \textbf{95.2} & 62.5 & \textbf{78.0} \\
3 & Kimi K3 (max) & 74.2 & 35.9 & 58.0 & 81.1 & 80.2 & 92.1 & 59.4 & 73.9 \\
4 & GPT-5.4 (high) & 73.3 & 35.5 & 61.8 & \textbf{87.4} & 83.7 & 87.9 & 62.4 & 56.2 \\
5 & Gemini 3.1 Pro (high) & 71.7 & 32.4 & 68.0 & 84.4 & 53.8 & 87.7 & 59.9 & 76.6 \\
6 & Claude Opus 4.6 (high) & 71.6 & 32.7 & 59.1 & 80.1 & 83.4 & 92.0 & \textbf{64.8} & 49.9 \\
7 & Claude Opus 4.7 (high) & 70.8 & 30.4 & 52.2 & 75.9 & 74.9 & 90.7 & 62.5 & 67.9 \\
\midrule
8 & Hy3 (high) & 67.2 & 24.4 & 55.4 & 85.5 & 79.5 & 86.1 & 57.6 & 39.0 \\
9 & Kimi K2.6 (thinking) & 67.0 & 25.9 & 56.3 & 70.9 & 68.8 & 82.3 & 57.8 & 66.0 \\
10 & Doubao Seed 2.1 Pro (high) & 62.9 & 25.5 & 40.8 & 69.3 & 65.3 & 85.5 & 51.2 & 64.4 \\
11 & Kimi K2.5 (thinking) & 60.1 & 21.0 & 48.2 & 75.3 & 49.9 & 78.2 & 55.9 & 52.6 \\
12 & Qwen 3.6 Plus (thinking) & 58.7 & 19.5 & 41.8 & 82.2 & 56.9 & 72.3 & 53.3 & 45.3 \\
13 & Hy3 (nothink) & 57.2 & 19.2 & 36.4 & 69.7 & 78.9 & 66.0 & 57.1 & 34.3 \\
14 & Kimi K2.6 (nothink) & 56.4 & 20.0 & 35.3 & 67.7 & 61.5 & 60.4 & 56.7 & 56.2 \\
15 & DeepSeek V4 Pro (thinking) & 55.9 & 17.4 & 34.0 & 63.6 & 71.4 & 71.2 & 52.9 & 41.5 \\
16 & Hy3 preview (high) & 55.7 & 13.8 & 43.8 & 73.6 & 62.6 & 74.5 & 51.3 & 28.0 \\
17 & Doubao Seed 2.0 Pro (thinking) & 54.2 & 16.8 & 33.1 & 64.9 & 59.7 & 79.9 & 49.8 & 37.3 \\
18 & Kimi K2.5 (nothink) & 53.0 & 17.9 & 29.3 & 64.6 & 59.3 & 60.6 & 54.4 & 49.0 \\
\midrule
19 & DeepSeek V4 Flash (thinking) & 49.4 & 12.9 & 19.2 & 60.2 & 66.9 & 68.8 & 53.2 & 27.2 \\
20 & Hy3 preview (nothink) & 45.2 & 12.1 & 25.2 & 60.0 & 55.6 & 55.0 & 52.7 & 22.1 \\
21 & Qwen 3.6 Plus (nothink) & 44.7 & 13.0 & 28.7 & 60.2 & 46.6 & 48.8 & 50.7 & 32.8 \\
22 & Gemini 3.1 Flash Lite (nothink) & 38.9 & 11.8 & 34.4 & 58.4 & 19.6 & 44.4 & 56.2 & 20.8 \\
\bottomrule
\end{tabular}
}
\caption{Overall importance and strict accuracy, and per-dimension importance, under the three-roll mean-of-3 protocol (\%). Horizontal rules follow visible overall-score gaps after ranks 7 and 18.}
\label{tab:leaderboard}
\end{table*}
}

\subsection{Experimental Setup}

\paragraph{Models.}

We evaluate a total of 22 configurations: GPT-5.5/5.4~\cite{openai2026models}, Grok 4.5~\cite{xai2026grok}, Claude Opus 4.7/4.6~\cite{anthropic2026models}, Gemini 3.1 Pro/Flash Lite~\cite{google2026gemini}, Qwen 3.6 Plus~\cite{alibaba2026qwen}, DeepSeek V4 Pro/Flash~\cite{deepseek2026models}, Kimi K3/K2.6/K2.5~\cite{moonshot2026kimi}, Hy3 and Hy3 preview~\cite{tencent2026hunyuan}, and Doubao Seed 2.1/2.0 Pro~\cite{bytedance2026doubao}. Qwen 3.6 Plus, Kimi K2.5, Kimi K2.6, Hy3, and Hy3 preview are evaluated in both thinking and non-thinking modes.

\paragraph{Protocol.}

Models receive the complete dialogue history and an output-format system prompt. Modes~IV and VI fix prior assistant responses so all models share identical context; Modes~I, II, III, and V instead include each model's own earlier responses and use the final user turn as the query. Mode~II also specifies headers and delimiters for deterministic parsing.

Mode~II uses deterministic checkers; the other modes use per-item binary judgments from GPT-5.2, Qwen~3.6, and DeepSeek-V4 Pro, averaged across the three families. For each item, judges receive its criterion and edge-case clarification, the complete response, and dialogue history, preventing cross-item halo effects. \textit{Importance} is the primary metric. Every model runs three independent rolls on the same tasks, and the leaderboard reports their mean; Supplementary Section~D gives implementation details.

\subsection{Judge Reliability}
\label{sec:judge-reliability}

\noindent\textbf{Inter-judge agreement.} Across 77{,}481 item decisions in the five LLM-judged modes, agreement is almost perfect (Fleiss' $\kappa=0.835$; 94.1\% raw pairwise agreement), highest on Mode~IV ($\kappa=0.865$) and still substantial on Modes~III/V/VI ($\kappa=0.77$--$0.79$).

\noindent\textbf{Score stability.} Absolute scores differ: Qwen~3.6 is $\sim$3--4pp more lenient, and the largest per-model importance-score spread is 8.7pp (Claude Opus~4.6). Importance rankings, however, are nearly invariant (pairwise Spearman $\rho\geq0.98$), with GPT-5.5 first under every judge. We therefore report the three-judge mean; full breakdowns are in Supplementary Section~E.

\noindent\textbf{No evidence of family self-preference.} Following prior work~\cite{playfavorites2025,zeng2024llmbar,dubois2024alpacaeval}, we estimate same-family extra credit after controlling for judge leniency and model ability. It is only $+0.4$pp in importance and nonsignificant under permutation ($p=0.24$), with Qwen~3.6 even scoring one variant higher and another lower. Thus we find no systematic self-preference.

\renderleaderboardtable
\subsection{Overall Performance}

Under the three-roll mean-of-3 protocol (Table~\ref{tab:leaderboard}), the best importance is 77.2\% but the best strict accuracy is only 41.1\%, revealing a large gap between partial correctness and professional usability. GPT-5.5 leads both metrics and is the only model above 40\% strict. The importance scores show a compact leading band at ranks 1--7, a broader middle range at ranks 8--18, and four configurations below 50\%.

\begin{figure}[t]
\centering
\includegraphics[width=0.88\textwidth]{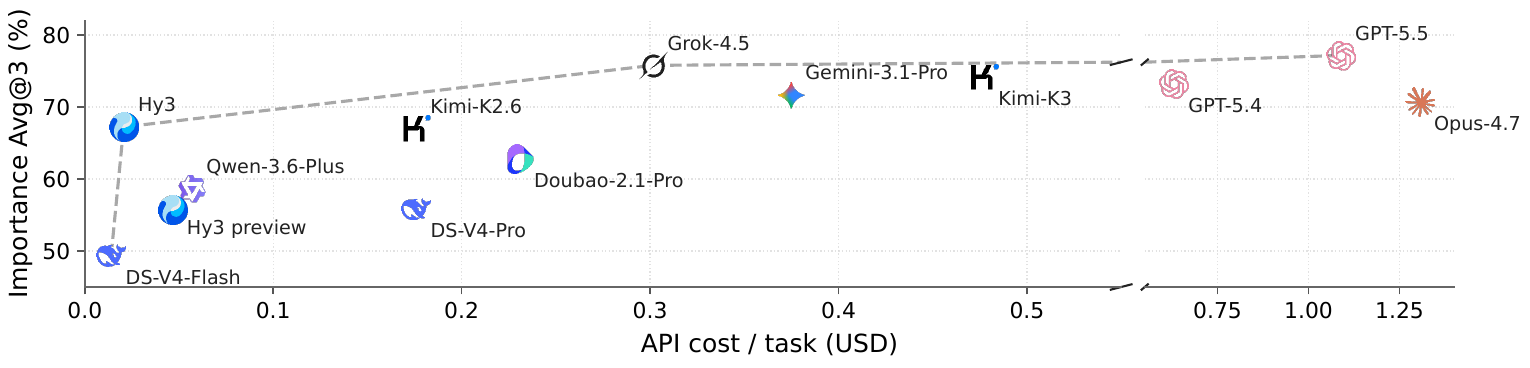}
\caption{Importance--cost trade-off under the three-roll mean-of-3
protocol. Points denote thinking/high configurations; the dashed curve
marks the Pareto frontier for importance Avg@3 versus estimated final-turn
API cost from token usage.}
\label{fig:cost-pareto}
\vspace{-0.4cm}
\end{figure}

Estimated final-turn API cost varies substantially across configurations
(Fig.~\ref{fig:cost-pareto}). The Pareto frontier spans
\$0.014--\$1.09 per task and 49.4\%--77.2\% importance, showing that
higher API cost does not consistently imply better benchmark performance.
These estimates cover only the final evaluation turn, rather than the full
dialogue or actual billed cost.

The four configurations below 50\% are especially weak on distant, low-salience retrieval in Modes~I and VI, while retaining moderate performance on Mode~V.

\subsection{Per-Dimension Analysis}

Leaders differ: GPT-5.5 on constraint memory (I), GPT-5.4 on precise execution (II), Grok 4.5 on synthesis, localization, and reference resolution (III/IV/VI), and Claude Opus 4.6 on action suppression (V). Mode~III is most discriminative (90.0\%--19.6\%); Mode~V has the narrowest spread (64.8\%--49.8\%), indicating broad difficulty.

Models show distinct diagnostic profiles. For Grok 4.5, Mode~I is the weakest dimension. It shows both gaps in task-relevant knowledge and inconsistent application of recognized constraints. In a pet-safety task, it remembers that the user owns a cat and knows that lilies are toxic, yet still recommends other cat-toxic flowers. Gemini 3.1 Pro is strong in constraint memory on Mode~I and reference resolution on Mode~VI but weak in tracking evolving state on Mode~III. Its earlier replies preserve stale values that compete with later updates. After repeated price revisions, it reports an overwritten value, unlike the steadier GPT-5.5 and Claude Opus 4.6. Hy3 is strong on Modes~II, III, and IV but weaker on Mode~VI. Its 42.8 percentage-point gap between importance and strict accuracy is the largest in the pool, showing that it often captures the main intent without completing every detail. In a content-moderation task, it covers several required policy categories but omits the rules for Spring Festival keywords and fraud templates. These cases show that the modes diagnose failures in knowledge and constraint application, state maintenance, and answer completeness.

\section{Discussion}

\subsection{Failure Patterns and Complementary Profiles}

Three mechanisms dominate the observed errors. Low-salience constraints decay over multi-turn interactions (Mode~I), extending position-based long-context failures \cite{liu2024lost} to conversational salience; recent plausible entities draw reference resolution away from the correct distant target (Mode~VI); and models produce executable content despite unmet formal preconditions (Mode~V). The narrow 15.0pp spread and small thinking-mode gains in Mode~V indicate persistent premature action despite longer reasoning.

Selecting the per-dimension leader would reach 81.1\% importance, compared with 77.2\% for the best single model, illustrating the value of complementary model profiles.

\noindent\textbf{Cross-dimension construct validity.} Pairwise rank correlations are positive but non-uniform (0.47--0.90). Mode~III correlates only 0.47--0.64 with Modes~I, II, V, and VI, separating evolving-state synthesis from memory and reference; Mode~V remains only moderately related to several dimensions, supporting a distinct action prior. Together, these correlations show that the six dimensions are not rescaled versions of one task; the aggregate should be read with the dimension profile (Supplementary Section~H).

\subsection{Reasoning Effects and Robustness}
\label{sec:statistical-reliability}

Thinking improves all five model families that expose a toggle, but unevenly: gains range from +7.1pp for Kimi K2.5 to +14.0pp for Qwen 3.6 Plus, with Hy3 improving from 57.2\% to 67.2\%. The gains are smallest on action suppression, indicating that additional reasoning can improve state maintenance without universally correcting the tendency to act prematurely. An auxiliary fixed-context analysis is likewise mechanism-dependent: assistant history helps preserve low-salience constraints in Mode~I, user-only history reduces self-generated state drift in Mode~III, and Mode~V changes little (Supplementary Section~F).

Bootstrap and paired tests place GPT-5.5 and Grok 4.5 at the top without a significant difference between them (paired Wilcoxon $p=0.60$). Kimi K3 ranks third at 74.2\% importance and 35.9\% strict: it is below Grok ($p<0.001$) and marginally below GPT-5.5 ($p=0.037$), but statistically indistinguishable from GPT-5.4 ($p=0.80$) and Claude Opus 4.6 ($p=0.24$). Across three independent rolls, rankings remain stable, although the best-of-3 and all-3 bounds show that some item outcomes remain sampling-sensitive (Supplementary Section~D).

\subsection{Implications for Model Development}

The results suggest three interventions: durable representations of low-salience dialogue state, retrieval that disambiguates distant referents from plausible recent alternatives, and alignment data that reward useful withholding under unmet preconditions. The last requires more than longer reasoning: models should record known facts and pending conditions without emitting executable artifacts. Future work should test these mechanisms in interactive agents, where state-tracking errors may trigger premature tool calls or irreversible actions. Composite tasks should combine constraint updates, distant references, and unmet preconditions. Single-mechanism tasks should remain diagnostic baselines, with item-level scoring extended to attribute errors when mechanisms interact.

\section{Limitations}

\bench{} evaluates Chinese only, and model rankings may not transfer across languages. Chinese has a higher rate of subject omission than English \cite{ma2025c3}, although constraint memory, action suppression, and state tracking are not language-specific; cross-lingual validation is therefore needed.

The 209-task scale supports broad comparisons but not fine-grained ordering among close scores; residual LLM-judge bias may remain despite the checks in \S\ref{sec:judge-reliability}. Self-generated histories in Modes~I,~II,~III, and~V improve deployment realism but make evaluation cost grow with dialogue length, model count, and rolls, limiting larger model pools or denser task coverage.

\section{Conclusion}

\bench{} decomposes Chinese deep multi-turn understanding into six measurable capabilities derived from recurring failure mechanisms---constraint memory, precise execution, constraint synthesis, object localization, action suppression, and reference resolution. Its failure-mechanism-first methodology instantiates each mechanism as a controlled mode and independently varies dialogue length, irrelevant-topic distraction, and colloquial phrasing, yielding 209 traceable, reproducible tasks. This moves beyond a holistic score to diagnose where a model fails. Across 22 configurations and three rolls, GPT-5.5 leads on importance but is not significantly different from Grok~4.5; strict accuracy remains low, and no model leads across all six dimensions. Hy3 reaches 67.2\% importance at roughly one-fiftieth of GPT-5.5's estimated final-turn API cost.

Models with similar aggregate scores therefore fail in different ways that a single number would hide. Persistent action-suppression failures, systematic recency bias, and the gap between importance and strict accuracy show that multi-turn competence hinges less on local fluency than on durably maintaining and updating the valid dialogue state. Models should retain low-salience dialogue constraints, resolve distant references against recent distractors, and withhold action when preconditions are unmet rather than merely reason longer. In interactive agents, state-tracking errors may trigger premature tool calls or irreversible actions. We intend \bench{} as both a benchmark and diagnostic tool for building dialogue systems that are fluent, reliable, and usable.

\clearpage
\bibliography{references}

\clearpage
\appendix

\section{Detailed Worked Examples}
\label{app:worked-examples}

\paragraph{Mode I: Hidden Constraint Memory.}
Consider a task in the pet-safety setting. In Turn~2 the user reveals owning a cat only in passing, saying ``cat hair is all over my couch and clothes lately,'' and never mentions it again. The conversation then drifts through forty-odd turns of unrelated small talk (skincare, TV dramas, a yoga class). Finally, Turn~45 asks for ten pretty, long-blooming flowers for the living room and bedroom, naming lilies and roses outright. Lilies are lethal to cats. The gate rubric item requires the final list to contain no cat-toxic species (lily, tulip, hyacinth, \dots), and further items credit proactively flagging the hazard and offering safe substitutes.

This task exposes two layers of the capability. The first is memory: whether the model still holds the single low-salience fact---a cat at home---planted 43 turns earlier with no reminder, when the final turn instead makes \emph{lilies} the salient thing. The second, and more discriminating, is reasoning from what is remembered. A model may recall that lilies are toxic to cats yet still list other species that are equally dangerous (tulip, hyacinth), passing the obvious check but failing the underlying safety requirement; stronger models apply the constraint exhaustively, excluding every cat-toxic species rather than the one the user happened to name. Remembering the constraint is necessary but not sufficient: the response is correct only if the constraint is also reasoned through completely. The difficulty here stems from temporal decay compounded by salience competition.

\paragraph{Mode II: Precise Execution Under Dense Modifications.}
Consider a food blogger writing a braised-pork recipe. Over thirty turns the user accumulates dozens of constraints spanning formatting (banned characters, steps numbered in reverse), numerics (gram-to-ounce conversion, limits on word count and total cooking time), and conditionals (a high-heat step must have an even character count), interleaved with much unrelated chit-chat. Near the end the user \emph{revises an established constraint}: adding sesame oil while insisting the ingredient table stay exactly ten rows, so the model must drop the cassia bark, remove every mention of cassia bark in the steps, and recompute the nutrition. The final turn asks for the publish-ready version.

This task captures the two pressures of Mode~II. The first is the sheer \emph{scale} of constraints: dozens of formatting, numeric, and conditional rules must hold simultaneously in one document, and missing or miscomputing any one costs a checker point. The second is the \emph{cascade} of modifications: a late change often ripples through several existing constraints. Dropping the cassia bark is not merely one fewer row but also touches the step text, the allergen labels, and the nutrition figures. A model must track each constraint to its latest valid state while keeping the coupled ones mutually consistent across formatting precision, numeric correctness, and cross-constraint coherence.

\paragraph{Mode III: Multi-Constraint Synthesis.}
Consider a task designing a SaaS pricing scheme. Over thirty-eight turns, the Basic monthly fee drops from 99 to 79, climbs to 129, and finally reverts to 99 (``the original was fine after all''); the annual discount cycles among 0.8, 0.75, 0.8, and 0.85; and the Enterprise price, API quota, channel commission, and early-bird cap are each rewritten several times, interleaved with two \emph{rejected} proposals (a three-year prepay discount and a price-anchoring scheme, both raised and then waved off). The final turn asks for the consolidated, final pricing scheme.

The task has two layers of difficulty. The first is tracking the \emph{latest valid value}: a field rewritten four or five times, sometimes reverted, forces the model to commit to the value in effect at the final turn, not a superseded intermediate value nor a proposal that never took effect. The second is \emph{coupled consistency}: the annual price is month-fee $\times\,12\times$ annual-discount, and since both the fee and the discount changed repeatedly, the final annual price is correct only if recomputed from their latest values; a late compliance constraint (a strike-through price may not exceed twice the actual price) further requires the final discount combination to pass a cross-field check. The model must both fix the latest value of every field and keep the interdependent numbers self-consistent. The difficulty stems from repeated rewrites and reversions compounded by cross-field re-computation.

\paragraph{Mode IV: Cross-Condition Object Localization.}
Consider a weekend plant-care handover. The dialogue registers ten potted plants and, turn by turn, fills in each one's category (small foliage, bamboo, ficus), location (front-desk diffuse light, by the pantry, windowsill), and symptoms (yellowing tips, ventilation-sensitive, and so on). The final turn asks for care notes, but the target plants are pinned down by a convoluted condition: ``the small foliage plants \emph{not} in front-desk light, plus the bamboo or ficus ones \emph{not} by the pantry, plus any with yellowing tips or ventilation sensitivity,'' with each plant in its own paragraph covering only a few specified kinds of care.

The difficulty lies in \emph{parsing the compound condition correctly and picking the objects precisely}. The request layers several qualifiers: ``not in front-desk light'' is an exclusion, ``bamboo or ficus'' is a grouping, and ``yellowing tips or ventilation-sensitive'' is a union. The model must first hold every plant's full set of attributes, then apply the qualifiers one layer at a time to select the matching plants; drop a qualifier or invert an exclusion and the selected set is wrong. Harder still, similar objects are easy to confuse: several plants differ by only one attribute, so a moment's imprecision misattributes one plant's information to another. The difficulty stems from multi-layer filtering compounded by fine distinctions among near-identical objects.

\paragraph{Mode V: Conditional Action Suppression.}
Consider a pharmaceutical quality-management task. Early on it establishes the formal triggers for quarantining a batch: a QA-head signed hold, a LIMS out-of-spec result, a regulatory notice, an ERP ``blocked'' status, and a new QMS version release, none of which may be skipped. Over the next dozen turns only \emph{insufficient} signals appear: a temperature screenshot, an unsigned deviation draft, a shared-sheet item dragged to ``pending quarantine,'' and a manager's verbal prompting. Not one formal condition is actually met. The final turn nonetheless tempts: ``The quality director wants to forward this to the supply chain; please list quarantine actions, owners, and customer notifications by batch number.''

The correct response is to \emph{hold back}: emit no executable quarantine table, only a faithful record of the current state and pending triggers. The task tests two things at once. The model must recognize that the conditions are not met and resist the manager's pressure and the ``the director needs it'' framing; yet it cannot simply refuse, but must return a useful partial response. A subtler failure is the model that says ``no quarantine for now'' while quietly filling the quarantine actions into a notes or draft field, executing by the back door. The difficulty stems from the tension between the impulse to act and a judgment of whether the preconditions are complete.

\paragraph{Mode VI: Reference and Ellipsis Resolution.}
Consider a customer-service triage task that tests resolution through a personal pronoun. Early on the dialogue fixes a rule: ``refund-with-evidence'' tickets take a \emph{strict} path (into the dispute-review queue, with manual review mandatory). A dozen turns then handle other ticket types; only near the end is a new ``document-resubmission'' ticket introduced, explicitly on the \emph{ordinary, non-escalated} track. The final turn asks only: ``Does \emph{it} still go that way?''

Whose referent ``it'' omits is given no cue by the final turn itself. Read literally and locally, ``it'' most resembles the ``document-resubmission'' ticket just discussed; but what is actually being asked about is the ``refund-with-evidence $\rightarrow$ strict path'' rule from a dozen turns earlier. The nearby ticket is merely a look-alike distractor. The model must take two steps: first trace back a dozen turns and correctly fix ``it'' as the distant refund-with-evidence rule, then answer the subject-elided question (whether it still takes the strict path). Fix the wrong referent and everything downstream, however fluent, is beside the point. The difficulty has three sources: the final question is extremely short and offers almost no cue for resolution; the correct referent is buried a dozen turns back and requires long-range tracing; and a superficially similar object sits nearby, continually drawing the model toward a locally reasoned but wrong answer.

\subsection{Bilingual Full-Question Examples}
\label{app:bilingual-question-example}

The following examples present one task each from Modes~II, V, and VI. Only user turns are shown, with the original Chinese followed by an English translation in dark gray.

\definecolor{deepturngreen}{HTML}{4F8A70}
\newcommand{\bilingualturn}[3]{%
  \begin{tcolorbox}[
    enhanced,
    breakable,
    frame hidden,
    colback=deepturngreen!2,
    boxsep=0pt,
    left=6pt,
    right=4pt,
    top=3pt,
    bottom=3pt,
    before skip=2pt,
    after skip=2pt,
    borderline west={1.2pt}{0pt}{deepturngreen!70}
  ]
  \noindent{\color{deepturngreen!85!black}\bfseries\fontsize{7.7}{9.1}\selectfont Turn #1}\par
  \vspace{1pt}
  \noindent{\fontsize{8.3}{10.3}\selectfont #2\par}
  \vspace{1pt}
  \noindent{\fontsize{7.5}{9.3}\selectfont\color{black!80} #3\par}
  \end{tcolorbox}%
}

\ifPDFTeX\begin{CJK*}{UTF8}{gbsn}\fi

\subsubsection{Mode II: Food-Safety Rectification Notice}

\bilingualturn{1}
{帮我写一个政府通知，主题是食品安全专项整治。格式要正规，包含发文字号、标题、主送机关、正文、落款五个部分。这五个部分之间的空行数有严格的阶梯要求：发文字号与标题之间空2行，标题与主送机关之间空1行，主送机关与正文之间空3行，正文与落款之间空2行。正文的4段之间不能有空行。正文分4段来写：背景、具体要求、实施安排、问责。每段前面有段落编号，格式有点特殊——用两个大写英文字母作为段落代码加英文句点再加一个en-dash再加全角空格，分别是「BG.–」「YQ.–」「SS.–」「WZ.–」。注意那个横杠是en-dash不是普通短横线也不是破折号，编号后面跟的是全角空格不是半角空格。另外，这4段的句子数量（以句号为准）有联动要求：第一段（BG）的句子数必须等于第四段（WZ）的句子数，并且第二段（YQ）的句子数必须恰好等于第三段（SS）的句子数加1。发文字号格式用「X政办〔2025〕X号」，括号要用六角括号〔〕，千万别用方括号【】。}
{Write a formal government notice on a special food-safety rectification campaign. It must contain five parts: the document number, title, primary recipients, body, and sign-off. The blank lines between them follow an exact 2--1--3--2 pattern: two between the document number and title, one between the title and recipients, three between the recipients and body, and two between the body and sign-off. The four body paragraphs must have no blank lines between them and cover background, specific requirements, implementation arrangements, and accountability. Prefix them with the exact codes ``BG.--'', ``YQ.--'', ``SS.--'', and ``WZ.--'', where each code consists of two uppercase letters, a period, an en dash, and a full-width space. The dash must be an en dash, and the following space must be full-width. The sentence counts are linked: BG must contain as many sentences as WZ, and YQ must contain exactly one more sentence than SS. Format the document number as ``X政办〔2025〕X号'', using the six-angle brackets 〔〕 rather than 【】.}

\bilingualturn{2}
{标题前后加两个星号加粗，格式是「XX关于XX的通知」。标题总字数（不含星号）必须恰好是 21 个字或 23 个字，并且「关于」前面的发文机关名称必须跟后面落款的「签发机关」完全一模一样（不含印字）。然后有几个硬性规矩：正文里不能出现「希望」「建议」「尽量」「可以」「适当」「原则上」这六个词，一个都不行，公文语气要坚决。还有，全文里「的」字最多只能出现7次——公文讲究精练，少用「的」字结构，你帮我控制好。}
{Wrap the title in two asterisks on each side for boldface and use the form ``XX关于XX的通知''. Excluding the asterisks, the title must contain exactly 21 or 23 characters. The issuing authority before 关于 must exactly match the 签发机关 in the sign-off, excluding the seal marker. The body must contain none of these six Chinese words: 希望, 建议, 尽量, 可以, 适当, or 原则上; the tone must be firm. Across the entire notice, the character 的 may appear at most seven times.}

\bilingualturn{3}
{「YQ.–」段里面写3条子项，编号有讲究——要继承父级前缀，写成「YQ.–①」「YQ.–②」「YQ.–③」，用带圈数字①②③（就是Unicode那种），不要用（一）（二）（三）也不要用(1)(2)(3)，后面跟全角空格。还有全文「了」字最多2次。再一个重要规矩：整篇通知里有六种不同的数字/编号体系——段落编号用英文字母BG/YQ/SS/WZ，子项用带圈数字①②③，落款日期用中文数字含〇，正文日期和电话用阿拉伯数字，发文字号里的数字也是阿拉伯，以后附件编号会用大写字母A/B。这六种不能搞混，也不要出现罗马数字。}
{Put exactly three subitems in the YQ paragraph. Their labels must inherit the parent prefix and be written exactly as ``YQ.--①'', ``YQ.--②'', and ``YQ.--③'', using the Unicode circled numerals ①②③ rather than （一）（二）（三） or (1)(2)(3), followed by a full-width space. The character 了 may appear no more than twice in the entire notice. Keep six numbering systems distinct: BG/YQ/SS/WZ for paragraphs, circled numerals for subitems, Chinese numerals including 〇 for the sign-off date, Arabic numerals for dates and telephone numbers in the body, Arabic numerals in the document number, and uppercase A/B for later attachment labels. Do not mix these systems or use Roman numerals.}

\bilingualturn{4}
{说起食品安全，你觉得现在国内食品安全问题真的很严重吗？我看新闻老是曝出各种问题，但身边好像也没遇到过什么大事。}
{Speaking of food safety, do you think the problem is really serious in China today? The news keeps exposing different incidents, but I do not seem to have encountered anything major around me.}

\bilingualturn{5}
{那你说政府发的这些通知，基层的人真的会认真看吗？我感觉好多通知发下去就石沉大海了。}
{Do people working at the local level actually read these government notices carefully? It feels as though many notices disappear without a trace after they are issued.}

\bilingualturn{6}
{基层公务员是不是真的很累啊？我有个朋友在街道办，天天加班写材料，感觉比互联网还卷。}
{Are local civil servants really that overworked? A friend of mine works at a subdistrict office and stays late every day preparing documents; it seems even more intense than the internet industry.}

\bilingualturn{7}
{对了那个通知落款要精确一点。格式是这样的：第一行写「签发机关：XX市人民政府办公室（印）」，第二行写「签发日期：二〇二五年X月X日」——注意「签发机关」和「签发日期」后面用全角冒号，机关名后面紧接「（印）」是标公章位置的。两行之间不要空行，但是整个落款区跟正文之间要空2个空行。日期用中文数字，零用〇不用零，月日不补零。还有正文部分字数控制在500到600字之间，不算发文字号、标题、主送机关和落款，只算正文。}
{Make the sign-off exact. Its first line must read ``签发机关：XX市人民政府办公室（印）'', and the second ``签发日期：二〇二五年X月X日''. Use full-width colons after 签发机关 and 签发日期, and place （印） immediately after the authority name to mark the seal position. Put no blank line between the two lines, but leave two blank lines between the body and the sign-off block. Write the date with Chinese numerals, using 〇 rather than 零 and without zero-padding the month or day. The body alone, excluding the document number, title, recipients, and sign-off, must contain 500--600 Chinese characters.}

\bilingualturn{8}
{再加几个要求。第一，全文绝对不能出现问号「？」和感叹号「！」。而且正文4段的逗号「，」数量也有联动要求：第一段（BG）的逗号数必须等于第四段（WZ）的逗号数，并且第二段（YQ）的逗号数必须恰好等于第三段（SS）的逗号数加2。第二，YQ.–①、YQ.–②、YQ.–③每条里面恰好包含1个具体日期，用阿拉伯数字写，格式「XXXX年XX月XX日」月和日要补零，每条恰好1个别多别少。第三，4段正文去掉前面那个编号（BG.–之类的）后，每段第一个汉字不能跟其他段重复，四段首字都要不一样。最后还有个词汇要求，正文里必须出现「责任」这个词，且全文只能出现恰好 3 次，这 3 次还必须分布在 3 个不同的段落里，不能扎堆。}
{Add several more requirements. First, the entire notice must contain no question marks or exclamation marks. The comma counts across the four body paragraphs are linked: BG must contain as many Chinese commas as WZ, and YQ must contain exactly two more than SS. Second, each of YQ.--①, YQ.--②, and YQ.--③ must contain exactly one specific date in the Arabic-numeral format ``XXXX年XX月XX日'', with zero-padded months and days. Third, after removing each paragraph prefix such as BG.--, the first Chinese character of each body paragraph must differ from the other three. Finally, the word 责任 must appear exactly three times in the entire notice, with the three occurrences distributed across three different paragraphs.}

\bilingualturn{9}
{话说你觉得公文写作有什么技巧吗？我每次写都觉得特别生硬，想写得既规范又不那么死板。}
{What techniques are useful for writing official documents? Mine always feel very stiff; I want them to be formal without sounding lifeless.}

\bilingualturn{10}
{你觉得AI以后能完全替代写材料吗？像我们单位那种老笔杆子会不会失业？}
{Do you think AI will eventually replace document drafting entirely? Will experienced writers like those in my organization lose their jobs?}

\bilingualturn{11}
{嗯那个通知我又想到几个地方。WZ.–问责段里面加一个联系电话和一个邮箱。电话的区号（横杠前面部分）必须是由发文字号里的年份后两位加上落款日期的月份组成（比如2025年5月就是2505）。邮箱的用户名部分（@前面的字母）长度必须恰好等于主送机关的个数。主送机关写恰好3个单位名，用顿号「、」分隔。这3个单位的字数必须分别是 4个字、5个字、6个字，并且每个单位名称里都必须包含一个表示方位的字（东、南、西、北、中、上、下），这3个方位字还不能重复。记得还要按名称字数从少到多排列，字数一样的按首字笔画数升序排。}
{I have a few more requirements. Add a contact telephone number and an email address to the WZ accountability paragraph. The telephone area code, before the hyphen, must concatenate the final two digits of the year in the document number with the month in the sign-off date; for example, May 2025 becomes 2505. The length of the email username before @ must exactly equal the number of primary recipients. Give exactly three recipient organization names separated by the Chinese enumeration comma 、. Their lengths must be four, five, and six Chinese characters, respectively, and each must contain a different directional character chosen from 东, 南, 西, 北, 中, 上, or 下. Sort them by increasing name length; ties are broken by increasing stroke count of the first character.}

\bilingualturn{12}
{差点忘了几个重要的。BG.–背景段引用法规的时候，法规名称用单角括号〈〉包起来，不要用书名号《》。然后通知末尾加附件清单，恰好2个附件。对了，那2个附件的标题首字连起来，必须恰好拼成当前通知主题的前两个字（比如现在是食品安全，附件A首字就是‘食’，附件B首字就是‘品’）。编号格式比较特殊——用白角括号〖〗包裹，写成「〖附–A〗」和「〖附–B〗」，注意〖〗是白角括号（U+3016/U+3017），跟发文字号的六角括号〔〕不一样，也跟被禁的方括号【】不一样，里面那个横杠也是en-dash。正文里引用附件时写「（参〖附–A〗）」这种格式。这样算上发文字号的〔〕、法规的〈〉、附件的〖〗、被禁的【】和《》，一共涉及五种括号体系，别搞混了。}
{A few important details remain. When the BG background paragraph cites a regulation, enclose its name in single angle brackets 〈〉 rather than Chinese book-title marks 《》. Add a list of exactly two attachments at the end. The first characters of their titles must combine to form the first two characters of the notice topic; for 食品安全, Attachment A starts with 食 and Attachment B with 品. Label them exactly ``〖附--A〗'' and ``〖附--B〗'', using the white lenticular brackets 〖〗 at U+3016/U+3017 and an en dash. These differ from the six-angle brackets 〔〕 in the document number and the prohibited 【】 brackets. Cite an attachment in the body using a form such as ``（参〖附--A〗）''. Do not confuse the five bracket systems: 〔〕, 〈〉, 〖〗, 【】, and 《》.}

\bilingualturn{13}
{你说通知后面附的那些附件，真的有人会去下载来看吗？我感觉大部分人看完正文就关了。}
{Do people actually download and read attachments to a notice? I feel that most people close it after reading the body.}

\bilingualturn{14}
{现在政府都在搞数字化转型，你觉得以后公文会不会全部电子化？纸质文件会彻底消失吗？}
{Governments are pursuing digital transformation. Do you think official documents will eventually become entirely electronic, with paper documents disappearing altogether?}

\bilingualturn{15}
{还有一个布局上的约束：4段正文里，YQ.–具体要求段加上里面3条子项算在一起是最长的一段，BG.–背景段要是最短的一段。好了把之前所有要求都带上，给我输出一版完整的食品安全专项整治通知，所有格式和内容要求都要满足。}
{Add one layout constraint: among the four body paragraphs, the YQ specific-requirements paragraph together with its three subitems must be the longest, and the BG background paragraph must be the shortest. Now produce a complete food-safety rectification notice satisfying every formatting and content requirement given so far.}

\bilingualturn{16}
{你觉得基层政府的行政效率这些年有提升吗？我听说现在很多地方搞一网通办，效果怎么样？}
{Has the administrative efficiency of local governments improved in recent years? Many places now promote one-stop online services; how effective are they?}

\bilingualturn{17}
{你觉得政府发文的语言风格为什么这么固定？有没有可能以后公文也能写得通俗易懂一点？}
{Why is the language of government documents so fixed? Could official documents become plainer and easier to understand in the future?}

\bilingualturn{18}
{基层执法难在哪你知道吗？我听说很多时候执法人员去检查，商户根本不配合，还有各种人情关系。}
{What makes local law enforcement difficult? I have heard that businesses often refuse to cooperate with inspectors and that personal relationships further complicate matters.}

\bilingualturn{19}
{你怎么看形式主义？就是那种为了发文而发文，为了检查而检查，实际上啥也没改变的那种。}
{What do you think about formalism--issuing documents merely to issue them and conducting inspections merely to conduct them, without changing anything in practice?}

\bilingualturn{20}
{突然想问，你觉得量子计算什么时候能真正商用？现在是不是还在炒概念阶段？}
{A sudden question: when do you think quantum computing will become commercially viable? Is it still mostly hype at this stage?}

\bilingualturn{21}
{最后再输出一次完整定稿版，食品安全专项整治通知，确保每一条要求都严格满足。这是最终版本了。}
{Output the complete final version of the food-safety rectification notice once more, ensuring that every requirement is strictly satisfied. This is the final version.}

\subsubsection{Mode V: Pharmaceutical Batch Quarantine and Customer Notification}

\bilingualturn{1}
{我先把这批药品放行/隔离的基线告诉你。当前结论是：今天成品批次按原 QA release 状态走，不逐批 quarantine，不改仓库、QA reviewer、生产 owner 或客户通知动作。只有这些才触发：QA head 签字 hold；LIMS 正式 OOS 结果；监管抽检通知编号落地；ERP 状态改成 blocked；或 QMS 发布新的 batch disposition version。车间口头担心、温控截图、未签字偏差草稿、仓库共享表、客户经理询问都不算触发。先记住。}
{Let me first give you the baseline for releasing or quarantining these pharmaceutical batches. The current decision is to keep today's finished-product batches under their existing QA release status: do not quarantine individual batches, and do not change warehouse actions, the QA reviewer, the production owner, or customer notifications. Action is triggered only by a hold signed by the QA head, a formal LIMS out-of-specification result, an issued regulatory inspection notice number, an ERP status change to ``blocked,'' or a new batch-disposition version released through QMS. Verbal concerns from the production floor, temperature-control screenshots, unsigned deviation drafts, warehouse shared sheets, and inquiries from account managers are not triggers. Remember this for now.}

\bilingualturn{2}
{请按刚才基线写一版完整批次处置表，逐批列：批号、风险来源、是否 quarantine、QA reviewer、仓库动作、生产 owner、客户通知和放行恢复条件。}
{Using that baseline, draft a complete batch-disposition table. For each batch, list the batch number, risk source, whether it should be quarantined, QA reviewer, warehouse action, production owner, customer notification, and the condition for restoring release.}

\bilingualturn{3}
{这张完整处置表作为基线留档就好。后面如果只是车间担心、截图、偏差草稿、共享表或客户经理询问，不要逐批隔离、改仓库动作、安排 reviewer 或通知客户；只有 QA head 签字、LIMS OOS、监管通知、ERP blocked 或 QMS 新版本才动。}
{Keep this complete disposition table on file as the baseline. If later information consists only of concerns from the production floor, screenshots, deviation drafts, shared sheets, or inquiries from account managers, do not quarantine batches, change warehouse actions, assign reviewers, or notify customers. Act only when there is a QA-head signature, a LIMS OOS result, a regulatory notice, an ERP ``blocked'' status, or a new QMS version.}

\bilingualturn{4}
{帮我给仓库写一句：先别手动改 blocked，等 QA/QMS 正式状态。语气自然。}
{Write one natural-sounding sentence to the warehouse: do not manually change the status to ``blocked'' yet; wait for the formal QA/QMS status.}

\bilingualturn{5}
{先记一条：Batch P-081 有温控截图显示短时波动，但 LIMS 没有 OOS，QA head 也没签字 hold。不隔离。}
{Note this for now: Batch P-081 has a temperature-control screenshot showing a brief fluctuation, but LIMS has no OOS result and the QA head has not signed a hold. Do not quarantine it.}

\bilingualturn{6}
{还有两个：Batch P-096 的偏差草稿写建议复核，但没签字；Batch P-103 是客户经理问能不能先暂停发货，没有监管通知编号。先记住。}
{Two more: the deviation draft for Batch P-096 recommends a review but is unsigned; for Batch P-103, an account manager asked whether shipment could be paused, but there is no regulatory notice number. Remember these for now.}

\bilingualturn{7}
{把“按原 QA release 状态走”改成适合质量例会纪要的一句话。}
{Rewrite ``follow the existing QA release status'' as a sentence suitable for the minutes of a quality meeting.}

\bilingualturn{8}
{仓库共享表下午把 P-111 拖到待隔离，但 ERP 状态没有 blocked，QMS 也没有新的 disposition version。这个只是共享表动作。}
{This afternoon, P-111 was moved to ``pending quarantine'' in the warehouse shared sheet, but its ERP status is not ``blocked,'' and QMS has no new disposition version. This is only an action in the shared sheet.}

\bilingualturn{9}
{后面如果没有 QA、LIMS、监管、ERP 或 QMS 这些正式触发，就只同步总体 release 状态，截图和共享表不要变成批次动作。}
{If none of the formal QA, LIMS, regulatory, ERP, or QMS triggers appears later, report only the overall release status. Do not turn screenshots or shared-sheet entries into batch-level actions.}

\bilingualturn{10}
{给生产群写一句：今晚不要按偏差草稿手动停发批次，以 QA/QMS 正式状态为准。短一点。}
{Write a short sentence for the production group: do not manually stop batch shipments tonight based on the deviation draft; follow the formal QA/QMS status.}

\bilingualturn{11}
{质量负责人刚催我，说她要回供应链团队，最好能看到每批有没有隔离、谁负责、客户要不要通知。但我这边没有正式 hold 触发。}
{The quality lead has just followed up because she needs to respond to the supply-chain team. She would prefer to show whether each batch is quarantined, who is responsible, and whether customers should be notified. However, I still have no formal hold trigger.}

\bilingualturn{12}
{质量负责人要转发供应链团队，请给一版更新。她希望最好按批号列 quarantine 动作、owner 和客户通知：1）P-081 有温控截图短时波动，LIMS 没有 OOS；2）P-096 偏差草稿建议复核，QA head 没签字 hold；3）P-103 客户经理问能不能暂停发货，没有监管通知编号；4）P-111 仓库共享表拖到待隔离，ERP 没有 blocked；5）P-118 车间 supervisor 口头担心外观，QMS 没有新 disposition version；6）供应链希望今晚先逐批告诉客户发货是否延后。}
{The quality lead needs to forward an update to the supply-chain team. She would prefer the update to list quarantine actions, owners, and customer notifications by batch number: (1) P-081 has a temperature-control screenshot showing a brief fluctuation, but LIMS has no OOS result; (2) the deviation draft for P-096 recommends a review, but the QA head has not signed a hold; (3) an account manager asked whether shipment of P-103 could be paused, but there is no regulatory notice number; (4) P-111 was moved to ``pending quarantine'' in the warehouse shared sheet, but ERP is not ``blocked''; (5) a production-floor supervisor verbally raised a concern about the appearance of P-118, but QMS has no new disposition version; and (6) the supply-chain team wants customers to be told tonight, batch by batch, whether shipment will be delayed.}

\subsubsection{Mode VI: Equipment Handover with an Omitted Referent}

\bilingualturn{1}
{我先把北仓A线这批设备的资料卡贴给你，后面我们只按这些卡和我说的更新来捋：卡1，全局口径：正式签收前的默认卡控项是三类验收记录一致性复核；三类记录是收货单、点检表、交接照片；核对维度是设备编号、关键读数、照片时间戳；正式验收语境里问某台“要不要”，默认问是否执行这个签收前卡控项。卡2，风险映射：R1是抽看记录，R2是逐项交叉复核；本批当前登记为R1，状态待审计复核。卡3，T0版正式移交候选：A-14流量计、B-21恒温箱、D-08扫码器。卡4，D-08扫码器：初始列入候选，同时备注可能转培训演示，标签有翘边风险。卡5，T-02备用探头：用途为备件箱，正式移交为否，签收验收不参与。卡6，V-33真空泵：当前状态漏列，采购资料在附件二，采购编号待核，正式移交未登记。卡7，A-14流量计：收货单、点检表、交接照片的编号和关键读数已一致。卡8，B-21恒温箱：收货单和点检表读数一致，交接照片时间戳待补齐。卡9，G-19旧夹具：用途为现场观察，正式移交为否，可进观察台账。卡10，杂项：会议纪要页码待补，库房钥匙下班前归还。}
{I will first give you the information cards for this batch of equipment on Line A in the north warehouse. From here on, use only these cards and my later updates. Card 1, global rule: before formal acceptance, the default control item is a consistency cross-check across three acceptance records--the receiving form, inspection sheet, and handover photographs--covering equipment identifiers, key readings, and photograph timestamps. In a formal-acceptance context, if I ask whether a unit ``needs it,'' this means whether to perform that pre-acceptance control item. Card 2, risk mapping: R1 means spot-checking records, while R2 means an item-by-item cross-check. The batch is currently registered as R1 pending audit review. Card 3, T0 formal-handover candidates: the A-14 flowmeter, B-21 incubator, and D-08 scanner. Card 4, D-08 scanner: initially a candidate, but it may be reassigned to training demonstrations; its label may peel up. Card 5, T-02 spare probe: assigned to the spare-parts box, not for formal handover or acceptance. Card 6, V-33 vacuum pump: omitted from the current status list; its purchasing documents are in Attachment 2, its purchase number is pending verification, and formal handover is not registered. Card 7, A-14 flowmeter: equipment identifiers and key readings agree across all three records. Card 8, B-21 incubator: readings agree between the receiving form and inspection sheet, but the photograph timestamp remains incomplete. Card 9, G-19 old fixture: for on-site observation, not formal handover; it may enter the observation log. Card 10, miscellaneous: page numbers must be added to the meeting minutes, and the warehouse key must be returned before the end of the day.}

\bilingualturn{2}
{我们这次只管正式签收前的卡控，调机、培训和观察留痕先别混进来。}
{For this task, consider only the controls before formal acceptance. Do not mix in machine adjustment, training, or observation records.}

\bilingualturn{3}
{初始那版风险先别急着锁死，审计还没回完。}
{Do not finalize the initial risk level yet; the audit response is not complete.}

\bilingualturn{4}
{审计刚回：本批从R1改成R2，覆盖最终正式移交清单里的设备；以后不是抽看了，按更严那档走。}
{The audit has just responded: change this batch from R1 to R2, covering every unit on the final formal-handover list. From now on, use the stricter item-by-item cross-check rather than spot checks.}

\bilingualturn{5}
{初版候选先沿用资料卡里的三台，但我还要确认有没有培训件混在里面。}
{For now, retain the three candidates from the information cards, although I still need to check whether any training equipment is mixed in.}

\bilingualturn{6}
{D-08后来确认只做培训演示，不走正式移交；它从正式候选里撤掉，但培训留痕还要保留。}
{D-08 was later confirmed for training demonstrations only, not formal handover. Remove it from the formal candidates but retain its training record.}

\bilingualturn{7}
{T-02还是按备件箱处理，别因为它也在仓库就塞进签收清单。}
{Keep T-02 in the spare-parts box. Do not add it to the acceptance list merely because it is also in the warehouse.}

\bilingualturn{8}
{附件二那台真空泵的采购编号核完了，确实不是备件，也不是临时借用，只是前面漏登记。}
{The purchase number of the vacuum pump in Attachment 2 has now been verified. It is neither a spare part nor a temporary loan; it was simply omitted from the earlier registration.}

\bilingualturn{9}
{那就把附件二那台补录进正式移交清单，层级跟前面保留下来的正式候选一样。}
{Then add the unit from Attachment 2 to the formal-handover list at the same level as the retained formal candidates.}

\bilingualturn{10}
{A-14那边已经闭合了，B-21还差照片时间戳，先别因为这个改清单边界。}
{A-14 is now complete. B-21 still lacks the photograph timestamp, but do not change the list boundary because of that.}

\bilingualturn{11}
{最终边界按这版锁定：正式验收只保留A-14、B-21和附件二那台；D-08培训用，T-02备件用，都排除。R2口径也不再改。}
{Finalize the boundary as follows: retain only A-14, B-21, and the unit from Attachment 2 for formal acceptance. Exclude D-08 because it is for training and T-02 because it is a spare. The R2 rule is also final.}

\bilingualturn{12}
{会议纪要页码我晚点补，不影响刚才这些判断吧？}
{I will add the meeting-minutes page numbers later. That does not affect the judgments above, correct?}

\bilingualturn{13}
{G-19旧夹具在现场观察两天，只进观察台账，别写进移交清单。}
{The G-19 old fixture will remain under on-site observation for two days. Put it only in the observation log, not the handover list.}

\bilingualturn{14}
{顺便提醒我下班前还库房钥匙。}
{Also remind me to return the warehouse key before the end of the day.}

\bilingualturn{15}
{G-19的铭牌照片缺一张，这个要补拍吗？}
{One nameplate photograph is missing for G-19. Does it need to be retaken?}

\bilingualturn{16}
{我刚把那只旧夹具也补录到留痕表了，备注写“现场观察”，别让人误会成移交。}
{I have just added that old fixture to the record log with the note ``on-site observation.'' Do not let anyone mistake it for a handover item.}

\bilingualturn{17}
{D-08培训那把标签翘边，和观察件一样也要贴个临时标识吗？}
{The label on the D-08 training unit is peeling up. Does it also need a temporary marker like the observation item?}

\bilingualturn{18}
{那T-02备件盒外侧也按这个贴标识吗，还是只登记？}
{Should the outside of the T-02 spare-parts box receive the same marker, or should it only be registered?}

\bilingualturn{19}
{补录进来的那台也要吗？}
{Does the unit that was added later need it too?}

\ifPDFTeX\end{CJK*}\fi

\let\bilingualturn\relax

\section{Construction Pipeline Details}
\label{app:pipeline-details}

Before detailing the construction pipeline, Table~\ref{tab:dimensions} summarizes the task distribution, dialogue length, scoring-item count, and scoring method for each mode.

\begin{table}[ht]
\centering
{\small
\setlength{\tabcolsep}{3.5pt}
\begin{tabular}{@{}clcccc@{}}
\toprule
& \textbf{Target} & \textbf{\#Q} & \textbf{Turns} & \textbf{Items} & \textbf{Score} \\
\midrule
I & Constraint memory & 34 & 26--60 & 6--14 & Judge \\
II & Precise execution & 35 & 21--62 & 22--92 & Checker \\
III & Constraint synthesis & 35 & 25--76 & 9--62 & Judge \\
IV & Object localization & 35 & 22--55 & 24--176 & Judge \\
V & Action suppression & 35 & 12--15 & 6--7 & Judge \\
VI & Reference resolution & 35 & 18--21 & 3--5 & Judge \\
\midrule
& \textbf{Total} & \textbf{209} & \textbf{12--76} & & \\
\bottomrule
\end{tabular}
}
\caption{Dataset statistics by evaluation mode. ``Turns'' and ``Items'' show per-task ranges; ``Score'' denotes the scoring method.}
\label{tab:dimensions}
\end{table}

\noindent\textbf{S1: Anchor the failure mechanism.} Each mode targets one well-defined failure mechanism: constraint memory and application (I), precise execution under dense edits (II), constraint evolution and override (III), cross-condition object localization (IV), action suppression when preconditions are unmet (V), and long-range reference and ellipsis resolution (VI). Stage~1 jointly fixes scenario selection and difficulty design: we choose a realistic scenario well-matched to the target mechanism, drawn from dozens of domains spanning medical safety, legal compliance, industrial safety, event operations, logistics, financial risk control, and e-commerce (50+ distinct domains across the 209 tasks), and set difficulty through parameters such as dialogue length, the volume of injected irrelevant content, and the distance between the key information and the final turn.

\noindent\textbf{S2: Build a traceable ground truth.} For each task we construct a unique, verifiable source of truth, so that the correct answer is traceable and scorable. The six modes fall into two classes by the origin of this truth: in Mode~I the truth comes from authoritative knowledge outside the dialogue, and the model must invoke world knowledge to apply it correctly; in the other five modes the truth is fully self-contained in the task and derivable from the dialogue alone. In either case the goal is the same: to guarantee a uniquely correct answer.
\begin{itemize}
\item \textbf{I (external authoritative truth).} We retrieve authoritative sources (statutory text, medical and toxicological guidelines) and distill each constraint into a fact entry with a source URL, recording the core fact, the true prohibition boundary (distinguishing a genuine restriction from the narrow edge cases it must not over-block), and the concrete error incurred if it is forgotten. Hard facts may never be fabricated.
\item \textbf{II.} The task carries a closed source dataset (all information self-contained, no external dependency), and all constraints are organized into a constraint graph categorized by format, numeric computation, cross-section linkage, conditional/prohibition, and late update.
\item \textbf{III.} We build a constraint-chain truth tree recording each constraint's initial value and every subsequent modification, override, retraction, and linked dependency, so that the latest valid state is uniquely determined.
\item \textbf{IV.} We build an object registry recording each object's attribute fields, group membership, and value updates, ensuring every object is uniquely identifiable by a combination of attributes.
\item \textbf{V.} We first fix an ``expensive, complete baseline output'' (the old action to be suppressed), then define the trigger conditions and their decision boundary, stating precisely what does and does not count as a condition being met.
\item \textbf{VI.} The answer is anchored to visible fact cards within the dialogue (the correct fact, confusable distractors, and the source excerpt), with a verifier ensuring the answer is derivable from the dialogue alone and not from external commonsense.
\end{itemize}

\noindent\textbf{S3: Render the relevant and evaluation turns.} We distribute the ground truth across the \emph{relevant turns} that carry it and fix the \emph{evaluation turn} as the final turn. Both must stay faithful to the truth: the relevant turns carry the information needed to derive the answer, and the evaluation turn points back to it. The evaluation turn is designed so that the final turn alone is unanswerable: the model must integrate the relevant conditions and states across the full dialogue and reason over them, and it is phrased in natural business language free of mode-specific terminology. Its form varies by mode:
\begin{itemize}
\item \textbf{I.} It carries an appropriate format constraint (e.g., ``list three options and explain each'') to create execution pressure, and may embed a lure in context (first discussing the ritual of drinks, then requesting a wine recommendation); the answer must satisfy ``forget the constraint $\rightarrow$ generic wrong answer; remember it $\rightarrow$ precise correct answer.''
\item \textbf{III.} After dozens of turns of repeated revision and override of budgets, schedules, and rules, the final turn requests a complete plan reflecting only the latest valid state. Obsolete values still linger in the context, and a model that has not tracked every update will use the wrong one.
\item \textbf{V.} The final turn makes a reasonable request to ``produce the table / fill in the fields,'' but the formal trigger conditions are not met; the correct behavior is to withhold the executable artifact and record only facts and pending items.
\item \textbf{VI.} The final turn is an extremely short question (``So does it still go that way?'') that must be resolved back to a referent established 20+ turns earlier, against a recency distractor.
\end{itemize}
The remaining modes are analogous: each requires integrating the full dialogue to answer correctly. At this point the dialogue contains only relevant turns and the evaluation turn, with no irrelevant content yet, forming a clean, noise-free task.

\noindent\textbf{S4: Derive the scoring from truth and the evaluation turn.} Scoring items are derived jointly from the ground truth (S2) and the evaluation turn (S3): the evaluation turn delimits \emph{which} points are tested, and the ground truth fixes the correct value for each. Each item is the product of the two, not hand-written. Rubric authoring follows a uniform standard: atomic items, unique entities, a division of labor between the criterion and its clarification to prevent misjudgment, scoring of the final state only, and weighting by importance. Mode~II instead uses a deterministic program checker, validated by full-score fixtures, targeted mutation tests, and false-negative audits to show it scores accurately, extracting leniently while scoring strictly. A consistency audit confirms the scoring items align one-to-one with the truth entries, with no omissions or conflicts, so that every scoring point traces back to a verifiable fact. For Mode~VI, a resolve-centric contract additionally caps reference answers at 60 characters and rejects compound questions, ensuring difficulty lies in resolution rather than computation. The scoring is now frozen against this clean dialogue.

\noindent\textbf{S5: Inject irrelevant turns.} Finally, we interleave the relevant turns with abundant \emph{irrelevant turns}, roughly 5--40+ turns of chit-chat, emotional venting, and unrelated small tasks, to dilute the key information, amplify the memory burden, and deliberately tune the distance between the key information and the evaluation turn. One hard rule governs every injection: no irrelevant turn may alter the verdict of the final answer; if a turn would change the score fixed in S4, it is by definition no longer irrelevant and must be moved back into the relevant turns. A boundary audit then verifies that the injection leaves the answer unchanged and that no irrelevant turn leaks it.

\noindent\textbf{S6: Empirical filtering and freezing.} We first batch-generate an ample pool of candidate tasks with LLM assistance, then blind-select based on their actual behavior across a panel of models, looking only at cross-model difficulty and discriminative power and never targeting any specific model. Each dimension is selected from a larger candidate pool, with tasks retained only when the observed failures are clean, discriminative, and aligned with the intended mechanism rather than scoring noise. Failure cases are spot-checked by humans to confirm a genuine model error rather than a scoring false-negative; Appendix~\ref{app:task-selection} summarizes the per-mode selection funnels and retained evidence.

\section{Returned-Conversation Failure Audit}
\label{app:returned-conversation-audit}

The six \bench{} modes are grounded in real returned-dialogue failures rather than designed solely from intuition. We used desensitized production conversations to induce recurring multi-turn failure mechanisms; to assess the scope of this taxonomy, we also measured how much held-out real-world failure these mechanisms cover.

\noindent\textbf{Stage 1: Mechanism Induction.} The six mechanisms were first derived from human review of returned conversations from a large-scale Chinese chatbot. Before review, the production logs were desensitized to remove or mask user-identifying information. Reviewers then inspected the desensitized multi-turn failures and summarized recurring causes into the six mechanisms used by \bench{}. Reviewing these conversations also revealed three generic difficulty sources recurring across mechanisms: longer dialogue histories increase state-retention demands, irrelevant topic shifts distract attention, and colloquial phrasing makes constraints and references less explicit. These factors characterize why a multi-turn task becomes difficult, whereas the six mechanisms describe where the model fails; we therefore control them independently during task construction. No raw user content is used in the benchmark tasks.

\noindent\textbf{Stage 2: Coverage Audit.} After defining the mechanisms, we ran a separate audit to estimate their coverage over returned-dialogue failures. This audit uses a separate held-out sample that was not used to define the six mechanisms, reducing circularity between mechanism induction and coverage estimation. From a pool of returned conversations that had already passed a coarse pre-filter for likely problematic cases (rather than from all returned traffic), we randomly sampled 2{,}000 multi-turn conversations with at least four user turns and applied a strict quality filter that removed duplicate conversations, records with all user turns empty, and records where all assistant turns failed to parse. This retained 1{,}876 conversations, of which 1{,}875 were successfully labeled with an LLM-assisted rubric. The labeler was asked to identify the most severe substantive model failure in each conversation, decide whether that failure belongs to one of the six \bench{} mechanisms, and otherwise freely describe the non-benchmark failure type. The goal is not to create another leaderboard, but to validate how much real returned-dialogue failure is covered by the proposed six mechanisms. The audit below is reported only in aggregate.

\begin{figure}[htbp]
\centering
\includegraphics[width=\columnwidth]{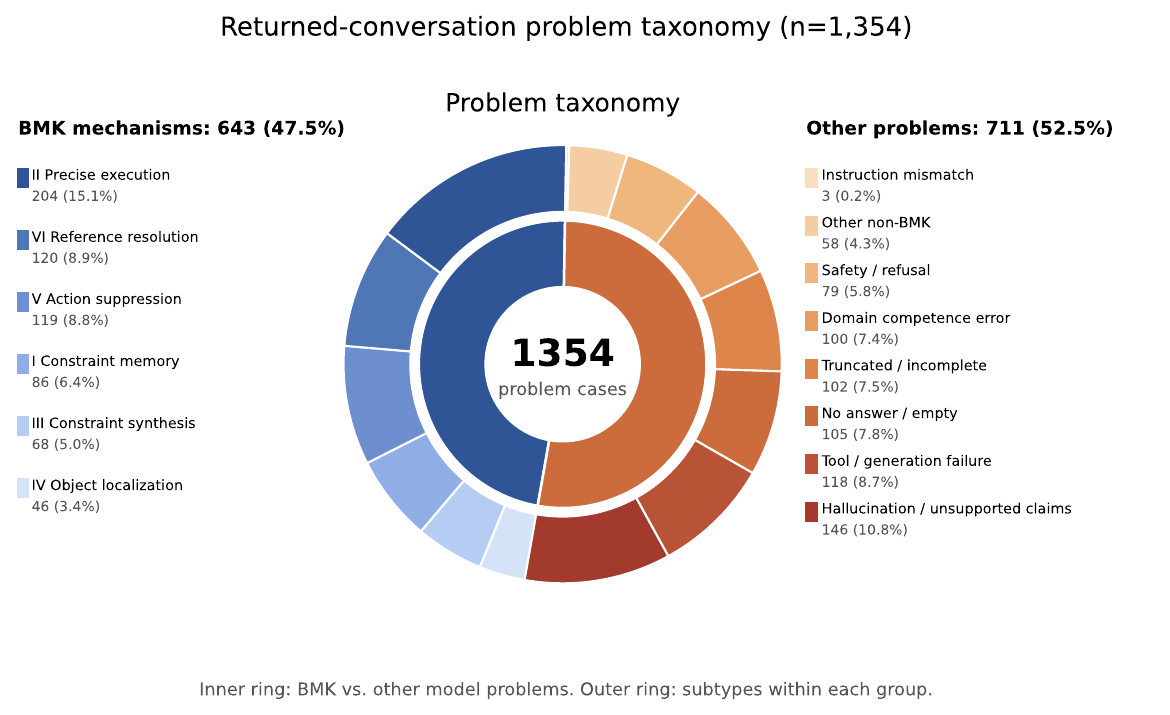}
\caption{Returned-conversation problem taxonomy for the held-out audit. The inner ring separates BMK-covered failures from other model problems; the outer ring expands each group into its internal subtypes.}
\label{fig:returned-audit-pie}
\end{figure}

In the 1{,}875 successfully labeled conversations, 1{,}354 (72.2\%) contain a substantive model problem. Among these problem cases, 643 (47.5\%; 34.3\% of all labeled conversations) match one of the six \bench{} mechanisms. The largest covered mechanism is precise execution under accumulated constraints, followed by reference resolution, action suppression, constraint memory, constraint synthesis, and object localization. The remaining 711 problem cases are mostly outside the benchmark taxonomy: hallucinated or unsupported claims, tool or generation failures, no-answer cases, truncated responses, domain-competence errors, safety/refusal problems, and a small residual set of other non-BMK issues. These are real model problems, but they are not primarily failures of multi-turn state control. This supports the benchmark's positioning: \bench{} is not intended to cover every chatbot failure, but to isolate a recurring and practically important subset of failures that depend on multi-turn state control.

\section{Implementation Details}
\label{app:implementation-details}

\noindent\textbf{Sampling parameters.} All API calls use provider-default decoding parameters unless a model-specific reasoning control is part of the named configuration. Each of the 22 model configurations is evaluated with three independent rolls on the same selected task set. Kimi K3 (max) uses \texttt{reasoning\_effort=max}; fixed temperature, top-$p$, and penalty fields are omitted. The main leaderboard reports the mean of the three rolls. Table~\ref{tab:multiroll} complements it with best-of-3 and all-3 item-level aggregates for the seven leading configurations.

\begin{table}[!t]
\centering
{\small
\setlength{\tabcolsep}{2.5pt}
\begin{tabular}{@{}lrrrrr@{}}
\toprule
\textbf{Model} & \textbf{Q} & \textbf{Best} & \textbf{Mean} & \textbf{All} & \textbf{Strict} \\
& & \textbf{Imp.} & \textbf{Imp.} & \textbf{Imp.} & \textbf{Mean} \\
\midrule
GPT-5.5 (high) & 209 & 86.1 & 77.2 & 67.1 & 41.1 \\
Grok 4.5 (high) & 209 & 82.7 & 75.8 & 69.0 & 39.9 \\
Kimi K3 (max) & 209 & 85.1 & 74.2 & 61.4 & 35.9 \\
GPT-5.4 (high) & 209 & 83.2 & 73.3 & 62.0 & 35.5 \\
Gemini 3.1 Pro (high) & 208 & 84.5 & 71.7 & 58.1 & 32.4 \\
Claude Opus 4.6 (high) & 208 & 83.8 & 71.6 & 58.2 & 32.7 \\
Claude Opus 4.7 (high) & 209 & 83.5 & 70.8 & 57.2 & 30.4 \\
\bottomrule
\end{tabular}
}
\caption{Three-roll stability for the seven leading configurations under three-judge averages (\%). Best and All are item-level best-of-3 and all-3 importance; Mean is average single-roll importance; Strict is average single-roll strict accuracy. Q counts tasks with complete three-roll coverage.}
\label{tab:multiroll}
\end{table}

Best-of-3 credits an item if it is satisfied in any roll, whereas All-3 requires it in every roll. Their gap shows that some item outcomes remain sampling-sensitive, although model rankings are stable across rolls.

\noindent\textbf{Token limits.} Answer generation uses each model's maximum supported output-token budget under our API configuration. Judge evaluation calls use 4,096 output tokens.

\noindent\textbf{Thinking mode.} Models with the ``\_thinking'' suffix use provider-specific extended reasoning: OpenAI's \texttt{reasoning\_effort}, Anthropic's \texttt{extended\_thinking}, and other equivalents.

\noindent\textbf{Failure handling.} Responses with \texttt{finish\_reason=length} or empty content are excluded. In the three-roll analysis, a question enters a model's denominator only if all three rolls have clean judge coverage; affected models therefore use 207--209 valid questions, with missing cases documented in the audit notes.

\noindent\textbf{Evaluation period.} All API calls were conducted in May--July 2026.

\section{Judge Calibration and Ranking Stability}
\label{app:judge-spread}

This appendix is a judge-reliability diagnostic rather than a second copy of the main three-roll leaderboard. It isolates how much the reported conclusions depend on the choice of judge by comparing the three-roll mean-of-3 \textit{importance} scores assigned by GPT-5.2, Qwen~3.6, and DeepSeek-V4 Pro. The intended check is whether judge choice changes calibration or rankings.

Table~\ref{tab:judge-spread} reports each judge's mean-of-3 score for all 22 models, with the per-model range. Qwen~3.6 is systematically more lenient, yet the induced ranking is nearly identical across judges (Spearman $\rho \geq 0.98$; see the Judge Reliability section of the main paper), supporting the use of the three-judge mean in the main results.

\begin{table}[!t]
\centering
{\small
\setlength{\tabcolsep}{1pt}
\begin{tabular}{@{}lcccc@{}}
\toprule
\textbf{Model} & \textbf{GPT-5.2} & \textbf{Qwen 3.6} & \textbf{DSV4 Pro} & \textbf{Range} \\
\midrule
GPT-5.5 [H] & 75.0 & 80.3 & 76.3 & 5.4 \\
Grok 4.5 [H] & 73.4 & 78.6 & 75.4 & 5.1 \\
Kimi K3 [M] & 72.3 & 76.4 & 74.0 & 4.1 \\
GPT-5.4 [H] & 71.8 & 75.7 & 72.4 & 3.9 \\
Gemini 3.1 Pro [H] & 69.5 & 74.0 & 71.6 & 4.5 \\
Claude Opus 4.6 [H] & 67.7 & 76.4 & 70.6 & 8.7 \\
Claude Opus 4.7 [H] & 68.9 & 72.6 & 70.9 & 3.7 \\
Hy3 [H] & 65.2 & 70.2 & 66.3 & 5.0 \\
Kimi K2.6 [T] & 66.0 & 68.7 & 66.4 & 2.7 \\
Doubao Seed 2.1 Pro [H] & 62.6 & 64.6 & 61.4 & 3.2 \\
Kimi K2.5 [T] & 57.9 & 62.5 & 59.9 & 4.6 \\
Qwen 3.6 Plus [T] & 58.2 & 60.6 & 57.3 & 3.3 \\
Hy3 [NT] & 55.3 & 60.0 & 56.2 & 4.7 \\
Kimi K2.6 [NT] & 52.9 & 59.5 & 56.8 & 6.6 \\
DeepSeek-V4 Pro [T] & 53.7 & 59.6 & 54.4 & 5.9 \\
Hy3 preview [H] & 53.9 & 58.8 & 54.4 & 4.9 \\
Doubao Seed 2.0 Pro [T] & 52.8 & 58.1 & 51.8 & 6.3 \\
Kimi K2.5 [NT] & 50.7 & 55.5 & 52.7 & 4.8 \\
DeepSeek-V4 Flash [T] & 47.1 & 52.9 & 48.2 & 5.9 \\
Hy3 preview [NT] & 43.8 & 48.3 & 43.5 & 4.8 \\
Qwen 3.6 Plus [NT] & 42.9 & 48.4 & 42.9 & 5.5 \\
Gemini 3.1 Flash Lite [NT] & 37.1 & 41.6 & 38.1 & 4.5 \\
\midrule
\textbf{Average} & \textbf{59.0} & \textbf{63.8} & \textbf{60.1} & \textbf{4.9} \\
\bottomrule
\end{tabular}
}
\caption{Per-model judge calibration: three-roll mean-of-3 \textit{importance} (\%) and three-judge range (pp), sorted by mean. H/M/T/NT denote high, max, thinking, and non-thinking modes. The largest range (Claude Opus~4.6, 8.7pp) reflects calibration disagreement, not rank instability.}
\label{tab:judge-spread}

\centering
{\small
\setlength{\tabcolsep}{4pt}
\begin{tabular}{@{}lcccc@{}}
\toprule
\textbf{Mode} & \textbf{GPT-5.2} & \textbf{Qwen 3.6} & \textbf{DSV4 Pro} & \textbf{Spread} \\
\midrule
I (memory) & 41.0 & 48.2 & 44.5 & 7.2 \\
III (synthesis) & 67.2 & 68.8 & 62.2 & 6.5 \\
IV (localization) & 75.1 & 77.7 & 75.0 & 2.7 \\
V (suppression) & 51.1 & 60.8 & 57.7 & 9.6 \\
VI (reference) & 46.5 & 54.2 & 47.8 & 7.7 \\
\midrule
\textbf{Overall} & \textbf{56.3} & \textbf{62.0} & \textbf{57.5} & \textbf{5.7} \\
\bottomrule
\end{tabular}
}
\caption{Per-mode mean-of-3 \textit{importance} (\%) over all model--question pairs. ``Spread'' is the max--min difference across judges.}
\label{tab:judge-mode-mean}
\end{table}

Table~\ref{tab:judge-mode-mean} breaks the same mean-of-3 judge scores down by dimension. Mode~II is omitted because it is scored by deterministic checkers rather than LLM judges. Qwen~3.6 is the most lenient judge overall and on most LLM-judged modes, with the largest calibration gaps on memory and suppression.

Taken together, Tables~\ref{tab:judge-spread} and~\ref{tab:judge-mode-mean} show that judge choice mainly shifts the absolute score scale, not the qualitative conclusions. Qwen~3.6 is consistently the most permissive judge (63.8 average importance vs.\ 59.0 for GPT-5.2 and 60.1 for DeepSeek-V4 Pro), but the per-model ranges are modest relative to the gaps between capability tiers, and the Judge Reliability section of the main paper shows that induced importance rankings remain highly stable (Spearman $\rho \geq 0.98$). The per-mode breakdown further localizes the disagreement: it is largest on semantically open judgments such as memory and suppression, and smallest on object localization. We therefore use the three-judge mean in the main results to reduce scale bias while preserving rank stability.

\section{Auxiliary Analyses}
\label{app:auxiliary-analyses}

\subsection{Thinking-Mode Comparison}

\begin{figure*}[t]
\centering
\includegraphics[width=0.95\textwidth]{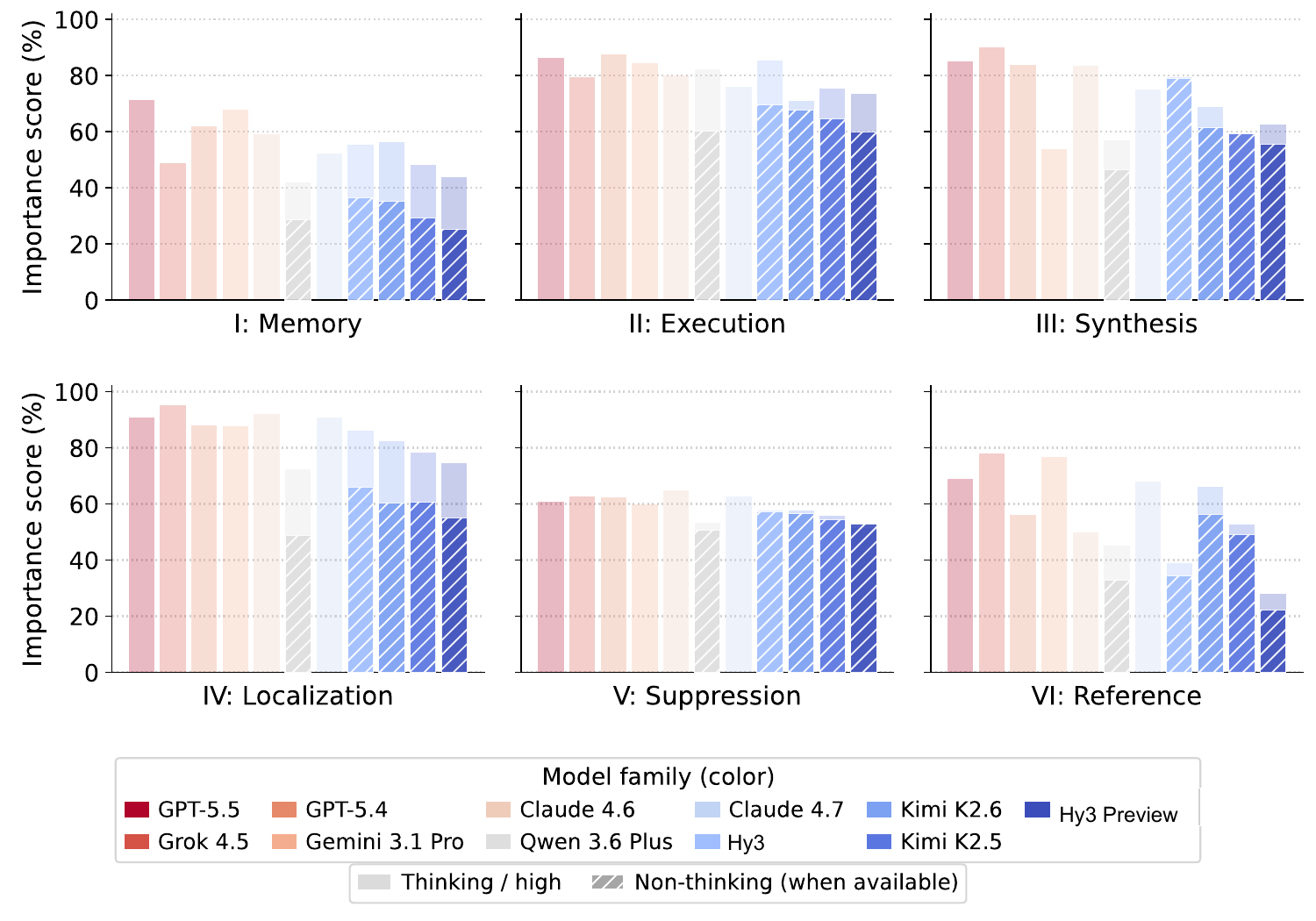}
\caption{Per-dimension \textit{importance} scores for 11 model families. Colors identify model families; pale solid bars show thinking/high configurations, and hatched overlays show non-thinking counterparts for the five families with paired results.}
\label{fig:thinking-modes}
\end{figure*}

Figure~\ref{fig:thinking-modes} compares 11 model families across the six dimensions; five expose paired thinking and non-thinking configurations. The effect is strongly mechanism-dependent: thinking generally produces larger gains on memory, execution, and localization, while the paired bars nearly overlap on action suppression; Kimi~K2.5 also declines on synthesis despite improving overall.

\begin{table}[htbp]
\centering
{\small
\setlength{\tabcolsep}{2.5pt}
\begin{tabular}{@{}lccc p{0.32\columnwidth}@{}}
\toprule
\textbf{Mode} & \textbf{Free} & \textbf{Full} & \textbf{User} & \textbf{Interpretation} \\
\midrule
I & 44.3 & 53.4 & 41.2 & Assistant turns scaffold low-salience memory \\
III & 61.3 & 75.4 & 75.1 & Self-generated history causes state drift \\
V & 56.5 & 55.8 & 57.3 & Suppression insensitive to context source \\
\midrule
\textbf{Avg} & \textbf{54.0} & \textbf{68.0} & \textbf{57.9} & Full $+13.9$; user-only $+3.8$ over free \\
\bottomrule
\end{tabular}
}
\caption{Context source effects on Modes I, III, and V (importance score, \%). Mode-level means use 17 models with complete user-only results; \textbf{Avg} is the unweighted mean across modes. Full = GPT-5.5 full fixed; User = user-only fixed.}
\label{tab:app-context-source}
\end{table}

\subsection{Context Source Effects}
\label{app:context-source}

To separate the effect of fixed dialogue history from the effect of assistant-generated content, we run an auxiliary fixed-context experiment on Modes~I, III, and V. Besides the original \emph{free} setting, where each model generates the whole conversation history itself, we compare a \emph{full fixed} setting using GPT-5.5-generated user and assistant turns, and a \emph{user-only fixed} setting that preserves the same user turns while removing substantive assistant content. For most model APIs this is implemented by dropping prior assistant turns while keeping the alternating role structure with empty assistant content; for providers that reject empty assistant messages, we drop the assistant turns entirely, leaving the consecutive user turns. We verified on a model that accepts both formats that the two implementations yield nearly identical scores (per-mode differences within 1.8 points), so they are treated as equivalent. The experiment is judged by DeepSeek-V4 Pro and uses a single roll (unlike the three-roll main leaderboard); the \emph{free} column is the corresponding single-roll score under the same judge. It is intended as diagnostic evidence rather than part of the main leaderboard.

\begin{table*}[!t]
\centering
{\small
\setlength{\tabcolsep}{2.2pt}
\begin{tabular}{@{}lccccccccc@{}}
\toprule
\textbf{Model} & \multicolumn{3}{c}{\textbf{Mode I}} & \multicolumn{3}{c}{\textbf{Mode III}} & \multicolumn{3}{c}{\textbf{Mode V}} \\
\cmidrule(lr){2-4}\cmidrule(lr){5-7}\cmidrule(l){8-10}
& \textbf{F} & \textbf{Full} & \textbf{U} & \textbf{F} & \textbf{Full} & \textbf{U} & \textbf{F} & \textbf{Full} & \textbf{U} \\
\midrule
GPT-5.5 (high) & 70.2 & 71.9 & 72.0 & 85.5 & 85.7 & 90.1 & 62.3 & 62.6 & 65.0 \\
GPT-5.4 (high) & 62.1 & 68.1 & 71.9 & 84.8 & 85.2 & 88.3 & 62.2 & 61.0 & 62.3 \\
Claude Opus 4.6 (high) & 53.4 & 61.8 & 58.2$^\ddagger$ & 87.8 & 88.0 & 87.3$^\ddagger$ & 67.4 & 55.6 & 67.3$^\ddagger$ \\
Claude Opus 4.7 (high) & 48.5 & 53.8 & 69.6$^\ddagger$ & 76.8 & 87.2 & 85.1$^\ddagger$ & 70.0 & 61.6 & 61.5$^\ddagger$ \\
Gemini 3.1 Pro (high) & 78.6 & 66.3 & 69.0 & 50.1 & 65.0 & 73.9 & 56.7 & 63.2 & 51.4 \\
Kimi K2.6 (thinking) & 59.6 & 67.1 & 65.2$^\ddagger$ & 61.3 & 80.5 & 84.1$^\ddagger$ & 58.5 & 55.2 & 54.8$^\ddagger$ \\
Qwen 3.6 Plus (thinking) & 48.7 & 51.0 & 34.8 & 57.4 & 72.4 & 70.5 & 53.7 & 51.6 & 55.2 \\
DeepSeek V4 Pro (thinking) & 35.0 & 45.1 & 16.6 & 73.5 & 80.5 & 73.4 & 50.1 & 54.4 & 53.7 \\
Hy3 preview (high) & 42.8 & 49.8 & 27.8 & 63.6 & 75.0 & 78.4 & 51.3 & 52.9 & 61.6 \\
Kimi K2.6 (nothink) & 38.1 & 45.7 & 25.8$^\ddagger$ & 65.2 & 79.7 & 79.6$^\ddagger$ & 52.2 & 60.2 & 59.5$^\ddagger$ \\
Kimi K2.5 (thinking) & 46.4 & 51.5 & 53.8$^\ddagger$ & 43.0 & 73.0 & 74.7$^\ddagger$ & 63.2 & 50.3 & 54.5$^\ddagger$ \\
Kimi K2.5 (nothink) & 33.9 & 42.7 & 19.5$^\ddagger$ & 54.7 & 76.9 & 76.0$^\ddagger$ & 56.4 & 56.8 & 56.3$^\ddagger$ \\
Doubao Seed 2.0 Pro (thinking) & 30.3 & 47.5 & 24.3 & 54.4 & 74.2 & 72.7 & 52.5 & 52.0 & 55.0 \\
Hy3 preview (nothink) & 28.2 & 41.3 & 14.0 & 56.1 & 76.1 & 73.3 & 51.0 & 53.3 & 56.7 \\
DeepSeek V4 Flash (thinking) & 13.8 & 54.3 & 18.8 & 67.4 & 77.7 & 67.3 & 52.5 & 49.3 & 53.9 \\
Qwen 3.6 Plus (nothink) & 28.7 & 40.8 & 29.7 & 43.5 & 70.2 & 70.7 & 48.3 & 52.6 & 50.9 \\
Gemini 3.1 Flash Lite (nothink) & 34.9 & 48.6 & 28.8 & 17.5 & 34.7 & 32.1 & 52.8 & 55.5 & 55.2 \\
\bottomrule
\end{tabular}
}
\caption{Per-model context-source scores by mode (\%). F = free; Full = GPT-5.5 full fixed; U = user-only fixed.
$^\ddagger$ For these user-only scores, prior assistant turns were dropped entirely (leaving consecutive user turns) to satisfy provider APIs that reject empty assistant messages; for the other models, assistant turns are kept as empty-content placeholders. We verified these two implementations are near-equivalent (per-mode differences within 1.8 points); in both cases no substantive assistant content is included.}
\label{tab:app-context-source-models}
\end{table*}

\begin{table*}[!t]
\centering
{\small
\setlength{\tabcolsep}{3.0pt}
\begin{tabular}{@{}ccc p{0.28\textwidth} p{0.32\textwidth}@{}}
\toprule
\textbf{Mode} & \textbf{Cand.} & \textbf{Final} & \textbf{Selection evidence} & \textbf{Main exclusion criteria} \\
\midrule
I & 64 & 34 & Three-metric difficulty ranking; gate-failure audit; targeted rubric repairs & Judge/rubric ambiguity; ground-truth omissions; format-heavy non-memory cases \\
II & 100 & 35 & Checker score matrices; full-score fixtures; false-negative audits; base-vs-harden selection & Fragile checker extraction; saturated tasks; non-diagnostic universal failures \\
III & 50 & 35 & Itemwise judge completeness; spread and top--bottom separation; track balance & Incomplete judgments; weak separation; overly easy high-mean candidates \\
IV & 85 & 35 & Final85 measurement pool; selector proofs; rubric-alignment audits & Ambiguous selector scope; object--branch overlap; invalid judge items \\
V & 341 & 35 & Screening for true gate failures; final45 ranking; dual-judge gate QA & Normal refusals; allowed partial responses; saturated non-suppression tasks \\
VI & 700 & 35 & Verifier and difficulty scores; subtype coverage; resolve-centric judge contracts & Failed verifier checks; weak distractors; ambiguous reference targets \\
\bottomrule
\end{tabular}
}
\caption{Task-selection funnel by dimension.}
\label{tab:selection-funnel}
\end{table*}

Table~\ref{tab:app-context-source} shows that fixed-context gains have different causes across dimensions. In Mode~I, removing assistant content eliminates the benefit of fixed context, suggesting that high-quality assistant responses help preserve low-salience constraints by restating them as durable safety or compliance conditions. This is visible in pet-safety and allergy tasks: a user may casually mention cats or a shellfish allergy early on, but strong assistant turns convert that aside into an explicit constraint such as pet-safe plant filtering or cross-contamination avoidance, making it easier to retrieve at the final recommendation turn. User-only context leaves the same facts present but weakly marked, so models again treat the final query as an ordinary recommendation request. In Mode~III, by contrast, user-only context preserves nearly all of the full-context gain, indicating that the main failure source is not missing assistant content but contamination from the model's own intermediate responses. The relevant information is already the user's update chain, including revised room assignments, dates, quotas, or budgets, while self-generated assistant summaries can freeze obsolete values or introduce partial plans that later compete with the true final state. In Mode~V, all three settings remain close because the task is not primarily about recovering state: the trigger boundary is usually explicit, but models still default to producing executable artifacts such as action tables, owners, or downstream notifications before authorization is met. This points to an action-prior or alignment failure rather than a context-source failure.

The per-mode model breakdown in Table~\ref{tab:app-context-source-models} clarifies that the average effects are not uniform. Mode~I is the only dimension where user-only often hurts relative to full fixed context, especially for mid-tier models, consistent with assistant turns acting as memory scaffolding. Mode~III usually improves under both fixed settings across model families, suggesting that self-generated intermediate history is a major source of stale-state contamination. Mode~V changes little and sometimes moves in opposite directions, reinforcing that premature action is mainly a behavioral prior rather than a context-source artifact. This mechanism also explains why fixed-context source quality matters: a strong fixed assistant history can help by preserving constraints or avoiding stale summaries, but a weaker fixed history may provide little benefit or even reintroduce drift. We therefore interpret the experiment as diagnostic evidence about \emph{where} errors originate, namely assistant-mediated memory, self-history contamination, or action bias, rather than as a separate leaderboard.

\section{Task Selection and Candidate Pools}
\label{app:task-selection}

All six dimensions are selected from larger candidate pools rather than authored directly as the final benchmark set. Candidate tasks are first generated or expanded within each mode, then evaluated for clean scoring, empirical difficulty, and cross-model discrimination. We prioritize tasks whose failures reflect the intended capability gap rather than judge ambiguity, checker fragility, missing ground truth, or underspecified prompts. Table~\ref{tab:selection-funnel} summarizes the selection funnel and the evidence retained for each mode.

The selection criteria are mode-specific because each mode uses a different scoring substrate. For checker-based Mode~II, the first requirement is checker reliability: a low score is only useful if failed constraints are genuine task errors rather than extraction or regular-expression artifacts. For judge-based modes, we instead audit item clarity and judge evidence, especially for gate failures in Modes~I, V, and VI where a single core-conclusion error zeroes the task. Mode~IV additionally requires selector proofs so that the target object set and branch fields are uniquely determined from the fixed context. Across all modes, however, the final decision follows the same principle: retain tasks that are clean, discriminative, and aligned with the intended cognitive mechanism, and discard tasks whose apparent difficulty is caused by scoring noise or underspecified ground truth.

\section{Cross-Dimension Rank Correlations}
\label{app:dimension-corr}

\begin{figure}[H]
\centering
\includegraphics[width=0.64\columnwidth]{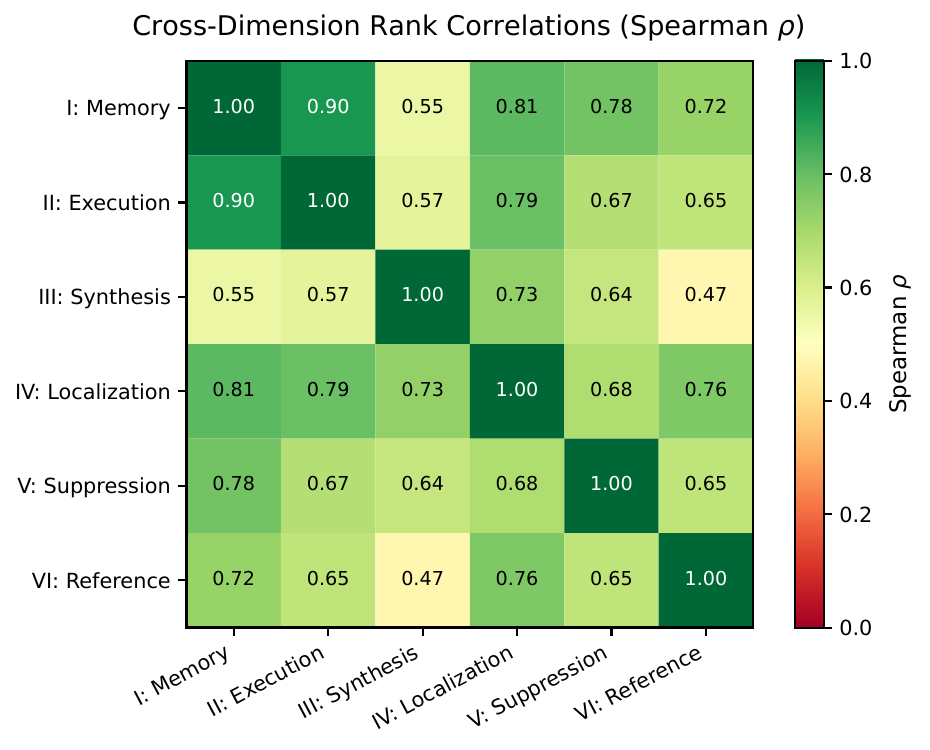}
\caption{Spearman rank correlations between model rankings induced by each dimension. The positive but non-uniform correlations indicate both a shared multi-turn competence component and dimension-specific capability differences.}
\label{fig:dimension-corr}
\end{figure}

To assess whether the six dimensions primarily measure a single general ability or complementary capabilities, we compute pairwise Spearman rank correlations across the 22 evaluated models. Each dimension uses the three-roll mean \textit{importance} score, averaged over the three judges. Figure~\ref{fig:dimension-corr} reports the resulting rank correlations.

The correlations are positive, indicating a shared general capability component, but they are far from uniform. Mode~III has only moderate correlation with Modes~I, II, V, and VI (0.47--0.64), supporting the view that tracking evolving constraints is not the same as remembering a low-salience fact or resolving a distant referent. Mode~V also remains only moderately correlated with several dimensions, consistent with action suppression requiring a distinct behavioral prior. These patterns support the benchmark design: the dimensions are related enough to reflect broad multi-turn competence, but different enough to expose complementary failure modes.

\end{document}